\documentclass[a4paper,11pt,final]{article}

\usepackage[colorlinks=true]{hyperref}
\usepackage[english]{babel}
\usepackage[utf8]{inputenc}
\usepackage[T1]{fontenc}
\usepackage{lmodern}
\usepackage[pdftex]{graphicx}
\usepackage[top=2cm, bottom=2cm, left=2cm, right=2cm]{geometry}
\usepackage{setspace}
\usepackage[french]{varioref}
\usepackage{lastpage}
\usepackage{fancyhdr}
\usepackage[table]{xcolor}
\usepackage{tikz}
\usepackage{titling}
\usepackage{graphicx}
\usepackage{amsmath}
\usepackage{amsthm}
\usepackage{amssymb}

\usepackage{multirow}
\usepackage{rotating}
\usepackage{subcaption} 
\usepackage[font=footnotesize,labelfont=bf]{caption}
\usepackage[gen]{eurosym}
\usepackage{dirtree}

\usepackage{rotating}
\usepackage{colortbl}
\usepackage{footmisc}

\usepackage{rotating}
\usepackage{threeparttable, tablefootnote}

\usepackage{diagbox} 

\usepackage{titlesec}
\titleformat{\paragraph}
{\normalfont\normalsize\bfseries}{\theparagraph}{1em}{}
\titlespacing*{\paragraph}
{0pt}{3.25ex plus 1ex minus .2ex}{1.5ex plus .2ex}

\usepackage{float}

\usepackage{tabularx}
\usepackage{enumitem}
\usepackage{longtable}
\usepackage{booktabs}

\newcommand{\fuda}[1]{%
    \begin{minipage}[t]{\linewidth}
    \begin{itemize}[nosep]
    #1%
    \end{itemize}
    \vspace{0.05cm}%
    \end{minipage}
}

\newcolumntype{C}[1]{>{\centering\let\newline\\\arraybackslash\hspace{0pt}}m{#1}}
\newcolumntype{L}[1]{>{\raggedright\let\newline\\\arraybackslash\hspace{0pt}}m{#1}}
\newcolumntype{R}[1]{>{\raggedleft\let\newline\\\arraybackslash\hspace{0pt}}m{#1}}

\usepackage{pifont} 
\newcommand{\cmark}{\text{\ding{51}}}
\newcommand{\xmark}{\text{\ding{55}}}

\newcommand\floor[1]{\lfloor#1\rfloor}

\usepackage{doi}
\usepackage[autostyle]{csquotes}
\usepackage[style=numeric, citestyle=numeric-comp, 
	maxcitenames=2, maxbibnames=99, 
    backend=biber
]{biblatex}
\usepackage
	{todonotes}

\usepackage{graphicx}

\definecolor{MYBLUE}{RGB}{55,126,184}
\definecolor{MYRED}{RGB}{228,26,28}
\definecolor{MYGREEN}{RGB}{77,175,74}
\definecolor{MYORANGE}{RGB}{255,145,0}
\definecolor{MYBROWN}{RGB}{205,133,63}

\renewcommand

\title{Characterizing and comparing external measures for the assessment of cluster analysis and community detection}
\author{Nejat Ar\i n\i k, Vincent Labatut \& Rosa Figueiredo}

\hypersetup{
    pdftitle={\thetitle},
    pdfauthor={\theauthor},
    bookmarksnumbered=true,bookmarksopen=true,
	unicode=true,colorlinks=true,linktoc=all,
	linkcolor=blue,citecolor=blue,filecolor=blue,urlcolor=blue,
	pdfstartview=FitH
}

\usepackage{enumitem}
\setlist{nolistsep}

\begin{document}
\maketitle
\sloppy

\abstract{In the context of cluster analysis and graph partitioning, many external evaluation measures have been proposed in the literature to compare two partitions of the same set. This makes the task of selecting the most appropriate measure for a given situation a challenge for the end user. However, this issue is overlooked in the literature. Researchers tend to follow tradition and use the standard measures of their field, although they often became standard only because previous researchers started consistently using them. In this work, we propose a new empirical evaluation framework to solve this issue, and help the end user selecting an appropriate measure for their application. For a collection of candidate measures, it first consists in describing their behavior by computing them for a generated dataset of partitions, obtained by applying a set of predefined parametric partition transformations. Second, our framework performs a regression analysis to characterize the measures in terms of how they are affected by these parameters and transformations. This allows both describing and comparing the measures. Our approach is not tied to any specific measure or application, so it can be applied to any situation. We illustrate its relevance by applying it to a selection of standard measures, and show how it can be put in practice through two concrete use cases.}


\textcolor{red}{\textbf{Cite as:} N. Arinik, V. Labatut \& R. Figueiredo. Characterizing and comparing external measures for the assessment of cluster analysis and community detection, in \textit{IEEE Access}. DOI: \href{https://www.doi.org/10.1109/ACCESS.2021.3054621}{10.1109/ACCESS.2021.3054621}}

\section{Introduction}
\label{sec:introduction}
\label{sec:Intro}

The problem of comparing two partitions of the same set occurs in a number of situations, the most widespread being probably the assessment of clustering (or cluster analysis) and community detection (or graph partitioning) results. In this context, one has computed the clusters of a dataset, or the community structure of a network. This result takes the form of a partition of the set of data points or of set of nodes, respectively. One then wants to compare this estimation with some ground-truth also taking the form of a partition. Alternatively, one has computed several such estimations, and wants to compare them to each other.

This comparison is traditionally performed through some measure able to quantify the similarity between two such partitions. In the context of cluster analysis, these are called \textit{external} measures, as they allow comparing the output of the clustering method to an independent solution (generally the ground truth). In the rest of this article, we will simply call them \textit{measures}, as there is no possible confusion in our context. Examples of such measures include \textit{Adjusted Rand Index} (ARI)~\cite{Hubert1985}, Normalized Mutual Information~\cite{Strehl2002} and so on. There are many ways to formalize what one means by "similar", resulting in the proposition of a very large number of measures over the years~\cite{Wagner2007, Meila2015}. In turn, this situation inevitably leads to the publication of a number of surveys aiming at reviewing and comparing all these measures~\cite{Vinh2010}.

In the literature, authors proposing new external measures follow a relatively standard workflow. First, they list some mathematical properties which they deem desirable in such measures, e.g. not being sensitive to the number of clusters $k$~\cite{Pfitzner2008, Rezaei2016, Fraenti2014}. They then show that existing measures do not possess these properties. Finally, they solve this issue by proposing a new measure having these properties, or modifying an existing one to this end.

There are mainly two ways to check whether a measure has a given property. The most robust approach is to proceed analytically, through a mathematical proof (e.g.~\cite{Meila2007}). 
However, this task requires certain skills, and can be difficult or even impossible depending on the considered measure and property. Moreover, the proof is generally not transposable to other measures and properties, which makes it a one-shot effort. This is why the second approach, which is empirical, is much more frequent in the literature (e.g.~\cite{Rabbany2013,Rezaei2016}). It consists in applying some predefined transformations to certain partitions, both designed in a way that is related to the property of interest, and to study how the measure reacts to these perturbations by using it to compare those partitions. For instance, to assess the sensitivity to $k$, one could increase the number of clusters in the transformed partition, and check how this affects the measure values.

Each application case is likely to bring its own constraints and requirements, so there is no such thing as a "best" measure that would fit \textit{all} situations. One trait considered as positive in one case could very well be perceived as a drawback in another. However, due to the profusion of available measures, selecting the most appropriate one for a given situation is a challenge for the end user. As mentioned before, some survey articles try to compare them, but they focus only a small number of measures~\cite{Vinh2010} and/or properties~\cite{Albatineh2006}. More importantly, the comparisons they perform are specific to these measures and properties~\cite{Vinh2010}, preventing the end user from including additional measures or properties in the comparison. In practice, the problem of selecting an appropriate measure to compare partitions is generally overlooked, and researchers tend to follow tradition and use the measures frequently appearing in the literature of their field.

In this work, we propose a new framework to solve this issue. It is based on the empirical approach mentioned above, and consequently relies on a set of predefined partitions and parametric partition transformations. We study the effect of each parameter on the measure through multiple linear regression, in order to produce results that the end user can interpret. Our framework is not tied to any specific measure, property, or transformation, so it can be applied to any situation. We illustrate its relevance by applying it to a selection of popular measures, and show how it can be put in practice through two concrete use cases. In addition to these contributions, we review the literature for desirable properties and the partition transformations used to test their presence, and propose a typology of the latter.

The rest of the article is organized as follows. 
First, in Section~\ref{sec:ExistingWorks}, we review the literature on external measures, focusing on desired properties, partition transformations, and property assessment methods. Next, in Section~\ref{sec:methods}, we introduce our own framework designed to study and compare measures and their properties. We put it into practice on a selection of widespread measures in Section~\ref{sec:Experiments} and discuss its results in Section~\ref{sec:Results}. Moreover, we consider two use cases in Section~\ref{sec:PracticalCases} to further illustrate its relevance. Finally, we review our main findings in Section~\ref{ref:Conclusion}, and identify some perspectives for our work.

\section{Literature Survey}
\label{sec:ExistingWorks}
In this section, we perform a review of the literature, focusing on three aspects directly related to our work. We first discuss the desirable properties used to characterize measures (Section~\ref{subsec:DesirableProperties}). We then survey the partition transformations proposed to empirically show the absence or presence of these properties (Section~\ref{subsec:ExistingTransformations}). Finally, we give an overview of the evaluation methods used for assessing and comparing the measures based on these transformations (Section~\ref{subsec:EvaluationApproaches}).

\subsection{Desirable Properties}
\label{subsec:DesirableProperties}
As mentioned in the introduction, measures can be characterized in terms of a number of distinct desirable properties. There are many of them, sometimes with minor differences, and the same property is likely to appear under different names and forms in the literature: this makes it difficult to list them exhaustively and compare them. Here, we focus on the most frequently used, and propose a typology to ease their comparison. These properties are listed in Table~\ref{tab:OverviewDesirableProperties}, with a short description, as well as examples of popular measures known to possess them. When the bibliographic sources explicitly name the property, we use the same name in the table. Otherwise, we propose a name based on its description. 
In the following, we distinguish three main categories, depending on whether the property is related to the \textit{measure interpretation}, to the way it handles \textit{random partitions}, and to its \textit{sensitivity} to certain characteristics of the partitions.

\subsubsection{Interpretation-Related Properties}
The first category of properties is related to the interpretability of a measure, i.e. how easily its values can be understood by a human operator. This concerns the interpretation of a single value, i.e. what its magnitude means, but also the comparison of several values, and the interpretation of their difference.

\textit{Understandability}~\cite{Dongen2000, Romano2014, Meila2015, OBrien2019} means that the measure has a straightforward interpretation. For instance, the Rand index~\cite{Rand1971} (RI) is the proportion of element pairs for which both partitions agree. Other measures have less direct interpretations, for example the Standardized Mutual Information~\cite{Romano2014} (SMI) is a normalized version of the mutual information corresponding to the number of standard deviations the mutual information is away from the mean value, for a specific null distribution. At the other end of the spectrum, composite measures such as the $F$-measure~\cite{Artiles2007} do not have a straightforward interpretation, as they combine other measures. This property is generally obtained by construction.

The \textit{Fixed Range} property~\cite{Wagner2007, Wu2009, Vinh2010} means that the measure is designed so that its values are restricted to a predefined interval, which is often $[0,1]$. This property eases the comparison of scores obtained on different partitions.

It is also the case of the \textit{Value Validity} property~\cite{Kvalseth2017}. Let $m_1$, $m_2$, $m_3$ and $m_4$ represent four numerical values obtained with some external measure for several pairs of partitions. In addition to order (size) comparisons such as $m_1 > m_2$, when a measure possesses the Value Validity property, the \textit{difference} between several pairs of partitions, such as $m_1 - m_2 > m_3 - m_4$ or $m_1 - m_2 > k (m_3 - m_4)$ (for constant $k$), can also be interpreted.

\textit{Convex Additivity}~\cite{Meila2003, Meila2007} concerns the case where one partition is a refinement of another partition (i.e. there is a hierarchical relationship between them). With a measure possessing this property, the difference in overall score can be expressed as a weighted sum of the score differences between individual clusters.


\subsubsection{Handling of Independent Partitions}
\label{subsubsec:HandlingIndependentPartitions}
The second category of properties focuses on how two independent partitions should be treated by the measure.

The \textit{Constant Baseline}~\cite{Wagner2007, Vinh2010, Romano2014, Rabbany2013, Albatineh2011} property deals with statistical independence, i.e. the case where one compares two partitions sampled independently at random. This property specifies that in this situation, the measure should return a constant value. In practice, this constant value is very often zero, in particular when the maximal value is $1$ (cf. also the \textit{Fixed Range} property), see for instance the Adjusted Rand Index~\cite{Hubert1985} (ARI).

The traditional approach to bring this property to an existing measure is to apply a so-called \textit{correction for chance}. It consists in subtracting to the measure the score estimated for two independent partitions, and possibly in normalizing the resulting expression, in order to get a fixed range. This is how Hubert \& Arabie derived their Adjusted Rand Index~\cite{Hubert1985} (ARI) from the original Rand Index~\cite{Rand1971}, but the method had been used before in other contexts~\cite{Goodman1954, Cohen1960}. Note that there is a number of ways to define the null model used to estimate the measure score under the assumption of independent partitions~\cite{Gates2017}, with no consensus emerging regarding which of these models is the most appropriate.

Certain authors consider two independent partitions as the worst possible case~\cite{Pfitzner2008}, meaning that the resulting score should correspond to the measure minimal value. On the contrary, others make a distinction between independence and worst case~\cite{Zhang2017, Gates2017b}, a property that is called \textit{Baseline-Minimum Distinction}. They generally place the constant baseline midway between the respective scores of the worst and base cases. This is for instance the case of the ARI, which ranges from $-1$ to $+1$, $0$ being the constant baseline. In practice though, cases with scores lower than the constant baseline are rare, and have not been studied much in the literature~\cite{Zhang2017}.


\subsubsection{Sensitivity to Partition Characteristics}
\label{subsubsec:sensitivity-partition-characteristics}
The last category of properties concerns the sensitivity of the measures to certain characteristics of the compared partitions. The main such characteristics are the number of clusters, the number of elements, the size of the clusters, and various descriptors allowing to express how similar the partitions are. These characteristics are often considered separately, and sometimes several at once.

In this category, the most frequent property is probably \textit{$k$-invariance}. Certain measures such as the Normalized Mutual Information tend to favor partitions depending on the number of clusters they contain when compared with a reference partition~\cite{Vinh2010}, a bias that a number of authors want to avoid~\cite{Wagner2007, Vinh2009a, Zhang2015, Romano2016, Amelio2016, Newman2020}. For example, suppose that one compares a ground truth partition to two estimated partitions differing only in their number of clusters. A biased measure will reach a noticeably higher value for one of these partitions due to this single difference.

By comparison, the \textit{Discriminativeness} property relies on the difference in number of clusters between the compared partitions~\cite{Rabbany2013, Amelio2016, Horta2015}. It states that the measure score should decrease when this difference increases. Put differently, the score should be larger when both partitions contain similar numbers of clusters than when they differ on this point.

The \textit{$n$-invariance} property is analogous to the $k$-invariance, except it is defined relative to the number of elements in the dataset~\cite{Meila2007, Wagner2007, Wu2009}, instead of the number of clusters. It allows comparing measure scores computed on datasets of different sizes, as $n$-invariant measures are not affected by such changes.

Authors do not agree on whether a measure should be sensitive or not to cluster size. This disagreement concerns partitions constituted of clusters which are imbalanced in terms of size, i.e. containing large and small clusters. Certain authors want the measure to focus mainly on the larger clusters, as they consider smaller ones as negligible~\cite{Wu2009, Hoef2019, Milligan1986}. Others adopt the \textit{Insensitivity to Cluster Size} property and assume that all clusters are equally important regardless of their size, and that the measure should not be sensitive to cluster size imbalance~\cite{Pfitzner2008, Rezaei2016, Fraenti2014}.

Finally, some properties focus on how the measure should quantify the differences between pairs of partitions. Suppose we compare one primary partition to two different secondary partitions, resulting in two scores. The \textit{Monotonicity} property states that the score of the most similar pair of partitions should be consistently higher or smaller (depending on whether the measure expresses similarity or dissimilarity)~\cite{Rezaei2016, Xiang2012, Gates2017b}. 
In addition, the \textit{Proportionality} property states that the difference between these scores should be proportional to how close the secondary partitions are~\cite{Liu2019}.
On the contrary, certain authors expect the measure score to rapidly change in presence of even the smallest differences, which corresponds to a non-linear behavior~\cite{Pfitzner2008}. 
More generally, some authors want the measure to be \textit{sensitive to small differences}~\cite{Fowlkes1983, Meila2007},
whereas some others, on the contrary, want the measure to ignore what are considered as marginal differences~\cite{Fraenti2014}.
It is important to stress that these are very generic properties, as the notion of proximity between two partitions can be understood in a number of ways.

\subsubsection{Discussion}
As explained in the introduction, and as summarized in Table~\ref{tab:OverviewDesirableProperties}, certain of the properties described in this section are obtained by construction, or verified through an analytical proof, whereas others are shown empirically, by applying specific transformations to a set of partitions. This is generally the case when the mathematical proof is impractical or too difficult to make. 

In this article, we adopt an empirical approach, therefore we focus only on the latter type of properties. This includes the properties of our second (Comparison with Random Partitions) and third (Sensitivity to Partition Characteristics) categories. The framework that we propose does not necessarily handles the properties exactly as they are described here: we sometimes had to reformulate them to ease experiments and make the framework more generic. It relies on a set of variables similar to those used in the literature to define these properties (number of clusters, number of elements, cluster size distribution, etc.). Our framework is able to handle properties on which authors disagree, such as the sensitivity to cluster size distribution or to small differences.

\begin{table}[ht!]
    \captionsetup{width=0.9\textwidth}
	\caption{Overview of the main desirable properties appearing in the literature, with examples of measures possessing them, and transformations used for their assessment.}
    \label{tab:OverviewDesirableProperties}
	\hspace{0.40cm}
	 \footnotesize
    \renewcommand{\arraystretch}{1.5}
    \begin{tabular}{p{1.75cm} | p{4cm} | p{2.75cm} | p{5.5cm}}
	    \hline
        \textbf{Category} & \textbf{Desirable Property} &  \textbf{Example measures} & \textbf{Related Transformations} \\
        \hline\hline
    	\multirow{4}{*}{\parbox[c]{1.5cm}{\centering Interpretation-Related}} 
            & Fixed Range~\cite{Wagner2007, Wu2009, Vinh2010} & NMI~\cite{Strehl2002}, NVI~\cite{Vinh2010} & 
                \fuda{ 
                    \item None (proof)
                }\\\cline{2-4}
            & Convex Additivity~\cite{Meila2007} & RI~\cite{Rand1971}, Mirkin~\cite{Mirkin1998}, VI~\cite{Meila2007}, $\chi^2$ distance~\cite{Hubert1985} &
                \fuda{ 
                    \item Splitting into unequal parts~\cite{Meila2007} (proof) 
                }\\\cline{2-4}
            & Understandability~\cite{Dongen2000, Romano2014, Meila2007, OBrien2019} 
                & RI~\cite{Meila2015}, JI~\cite{Meila2015}, SMI~\cite{Romano2014}, Split-Join~\cite{Dongen2000}
                & 
                \fuda{ 
                    \item None (proof)
                }\\\cline{2-4}
            & Value Validity~\cite{Kvalseth2017} & MI$_c$~\cite{Kvalseth2017} 
            & 
                \fuda{ 
                    \item None (proof)
                }\\
        \hline
        \hline
        \multirow{2}{*}{\parbox[c]{1.5cm}{\centering Handling of Independent Partitions}}
            & Constant Baseline ~\cite{Wagner2007, Vinh2010, Romano2014, Rabbany2013, Albatineh2011} 
                & ARI~\cite{Hubert1985}, AMI~\cite{Vinh2009a}, rNMI~\cite{Zhang2015}, RMI~\cite{Newman2020}, FNMI~\cite{Amelio2016}, cNMI~\cite{Lai2016}  & 
                \fuda{
                    \vspace{-0.2cm}
                    \item Fragmenting every cluster~\cite{Pfitzner2008}
                    \item Random shuffling~\cite{Vinh2009a, Romano2014, Rezaei2016, Romano2016, Amelio2016, Gates2017, Gates2017b}
                }\\\cline{2-4}
            & Baseline-Minimum Distinction~\cite{Zhang2017, Gates2017b}
                & ARI~\cite{Hubert1985}, SMI~\cite{Romano2014}  & 
                \fuda{
                    \vspace{-0.2cm}
                    \item Random shuffling~\cite{Gates2017b}
                }\\
        \hline
        \hline
        \multirow{7}{*}{\parbox[c]{1.5cm}{\centering Sensitivity to Partition Characteristics}}
            & $k$-invariance~\cite{Wagner2007, Vinh2009a, Zhang2015, Romano2016, Amelio2016, Newman2020}
                & ARI~\cite{Hubert1985}, VI~\cite{Meila2007} 
                & \fuda{
                    \vspace{-0.2cm}
                    \item Random shuffling~\cite{Vinh2009a, Zhang2015, Romano2016, Amelio2016, Gates2017b}
                    \item Splitting into singleton clusters~\cite{Newman2020} 
                    \item Swap with single cluster~\cite{Rezaei2016} 
                }\\\cline{2-4}
            & $n$-invariance~\cite{Meila2007, Wagner2007, Wu2009}
                & VI~\cite{Meila2007}, FMI~\cite{Fowlkes1983}, NMI~\cite{Strehl2002}, ARI~\cite{Hubert1985}  & 
                \fuda{ 
                    \item None (proof) 
                }\\\cline{2-4}
            & Discriminativeness~\cite{Rabbany2013, Amelio2016, Horta2015}
                & ARI~\cite{Hubert1985}, GNMI~\cite{Amelio2016}, FNMI~\cite{Amelio2016}  &
                \fuda{   
                    \vspace{-0.1cm}
                    \item Merging whole clusters~\cite{Rabbany2013, Amelio2016} \& Splitting into unequal parts~\cite{Rabbany2013, Amelio2016} 
                }\\\cline{2-4}
            & Sensitivity to Small Differences~\cite{Pfitzner2008, Fowlkes1983, Meila2007, Fraenti2014}
                & VI~\cite{Meila2007},FMI~\cite{Fowlkes1983}  & 
                \fuda{  
                    \item Swap with all clusters~\cite{Pfitzner2008} 
                }\\\cline{2-4}
             & Insensitivity to Cluster Size~\cite{Pfitzner2008, Rezaei2016, Fraenti2014}
                & PSI~\cite{Rezaei2016} &
                \fuda{
                    \item Swap with single cluster \& remove~\cite{Rezaei2016} 
                    \item Fragmenting a single cluster~\cite{Rezaei2016} 
                    \item Random shuffling~\cite{Hoef2019, Warrens2019}
                }\\\cline{2-4}
            & Monotonicity~\cite{Rezaei2016, Xiang2012, Gates2017b}
                & PSI~\cite{Rezaei2016}, Element-centric~\cite{Gates2017b}
                &
                \fuda{  
                    \vspace{-0.3cm}
                    \item Merging a whole cluster with a part of other cluster~\cite{Rezaei2016}
                    \item Merging parts of different clusters~\cite{Dom2002, Rosenberg2007}
                    \item Merging whole clusters \& splitting into equal parts~\cite{Xiang2012}
                    \item Swap with single cluster~\cite{Rezaei2016}
                    \item Swap with all clusters~\cite{Rezaei2016}
                    \item Random shuffling~\cite{Gates2017b}
                }\\\cline{2-4}
            & Proportionality~\cite{Meila2007, Liu2019} vs. Non-linearity~\cite{Pfitzner2008}
                & Kappa index~\cite{Liu2019} & 
                \fuda{  
                    \item Random shuffling~\cite{Liu2019}
                }\\ 
        \hline
	\end{tabular}\\
\end{table}

\subsection{Partition Transformations}
\label{subsec:ExistingTransformations}
Like for the desired properties, the literature exhibits a large number of different partition transformations, which are not always named, and when they are, not always similarly. This makes it difficult to identify and compare them. Here, we focus on the most frequent ones and use their most consensual names. Table~\ref{tab:OverviewDesirableProperties} indicates the transformations used in the literature to assess the presence of each listed property. One can distinguish two types of partition transformations: random vs. deterministic.

\subsubsection{Random Transformations}
Random transformations consist in randomly distributing all the elements of the reference partition over a number of clusters to form the new partition. These transformations mainly differ in the probability distributions they rely upon. Such processes can be seen more as shuffling than transformations, as the original partition has no effect on the result. In essence, the goal is to obtain a partition as independent as possible from the original one. They are mainly used to check the existence of the Handling of Independent Partitions category of properties~\cite{Vinh2009a, Romano2014, Rezaei2016, Romano2016, Amelio2016, Gates2017, Gates2017b}. 
But several works leverage random transformations to look for other desirable properties, too. Certain authors force the shuffled partition to have various numbers of clusters and imbalanced cluster sizes, in order to check the $k$-invariance~\cite{Vinh2009a, Zhang2015, Romano2016, Amelio2016, Gates2017b} and Insensitivity to Cluster Size~\cite{Rezaei2016,Warrens2019} properties, respectively. Others shuffle the original partition with an increasing level of randomness in order to test for the Monotonicity~\cite{Gates2017b} and Proportionality~\cite{Liu2019} properties.

\subsubsection{Deterministic Transformations}
Deterministic transformations are used more frequently in the literature, probably because they offer a better control of the changes applied to the original partition. We distinguish five categories of such transformations. We call the first one \textit{Remove}, and it consists in deleting some elements from a cluster without erasing it completely. Although it is used to check the Insensitivity to Cluster Size property in the literature~\cite{Rezaei2016}, it has the drawback of affecting simultaneously two aspects of the partition: cluster size, and number of elements $n$. For this reason, it is not frequently used.

The second transformation category is \textit{Split}, which consists in dividing a cluster into multiple smaller parts. Two variants mainly appear in the literature: splitting into \textit{equal}~\cite{Meila2007, Meila2015} vs. \textit{unequal} parts~\cite{Meila2007, Meila2015,Rabbany2013, Amelio2016}. There is also a specific case of the first variant, consisting in splitting a cluster into only singleton clusters~\cite{Rand1971, Reichart2009, Newman2020}. 
This transformation category is used in the literature to test several distinct properties. Hierarchical splits (i.e. refinements of a partition) constitute an important part of the small experiments proposed by Meil{\u{a}}~\cite{Meila2007, Meila2015}, and allow to check the Convex Additivity property. Reichart and Rappoport~\cite{Reichart2009} compare a reference partition to two estimated partitions differing mainly in their number of clusters: slightly perturbed reference vs. singleton clusters. They expect that singleton clusters are less similar to the reference, and a measure should not favor singleton clusters in such a case (cf. $k$-invariance property). Rabbany et al~\cite{Rabbany2013} apply repeated split operations onto the ground truth of several real-world networks and then compare them to check the Discriminativeness property.

Transformation \textit{Merge} is the reciprocal of Split, as it gathers nodes belonging to different clusters into the same cluster. It also appears under three forms: merging a whole cluster with a \textit{whole} other cluster~\cite{Meila2007, Rabbany2013, Amelio2016, Xiang2012} vs. a \textit{part} of another cluster~\cite{Rezaei2016}, and merging parts of different clusters~\cite{Rosenberg2007}. Note that the last two transformations are not \textit{pure}, in the sense that a Split is performed before the Merge. Regarding the desirable properties, since Merge is the reciprocal of Split, all the properties tested through Split can be also be tested by using Merge. On top of that, some authors leverage Merge to test for \textit{Monotonicity}, in two different ways: Rezaei and Fränti~\cite{Rezaei2016} enlarge incrementally a specific cluster by moving elements from the other clusters, whereas Rosenberg and Hirschberg~\cite{Rosenberg2007} merge same-sized parts of each cluster to create new clusters, which they consider as \textit{noise}.

The next two transformations can be viewed as combinations of Split and Merge, and they are also frequently used in the literature. \textit{Swap} consists in interchanging a number of elements between pairs of (generally equal-sized) clusters. In practice, this operation is usually repeated for each cluster, using one of two different forms: each cluster swaps elements with \textit{only one} different cluster~\cite{Meila2007, Rezaei2016} vs. \textit{all} other clusters~\cite{Meila2007, Pfitzner2008, Rezaei2016}. In the literature, the first form is mainly used with a range of the number of clusters to check the $k$-invariance property. In the experiments of Rezaei and Fränti~\cite{Rezaei2016}, the authors keep the cluster sizes fixed, independently from the number of clusters. However, this increases the number of total elements, which arguably introduces a side effect in their experiments. The second form induces more perturbation of the original partition compared to first one, and the experiments in the literature mainly focus on the desirable properties related to this aspect, which are Monotonicity~\cite{Rezaei2016} and Sensitivity to Small Differences~\cite{Pfitzner2008}.

Finally, the idea behind the \textit{Fragment} transformation is that elements belonging to the same cluster in the original partition are placed in different clusters in the transformed partition, as much as possible. Two variants mainly appear in the literature: fragmenting a \textit{single} cluster vs. \textit{all} of them. The former~\cite{Rezaei2016} 
is only used to change marginally the underlying partition structure, whereas the aim of the latter~\cite{Rand1971, Pfitzner2008, Hoef2019}
is to obtain two maximally different partitions.
In the literature, these variants are used to check the Insensitivity to Cluster Size~\cite{Rezaei2016, Hoef2019} and Constant Baseline~\cite{Pfitzner2008} properties, respectively.

\subsubsection{Discussion}
Besides these categories, the literature also contains transformations which can be expressed as combinations of some of these categories~\cite{Xiang2012, Rabbany2013, Rezaei2016}. It is important to stress that transformations are typically defined \textit{ad hoc}, specifically to test for a particular property of interest, and on some predefined partitions. For this reason, each author adopts a different angle, and it is hard to find two articles with the exact same methodology, targeting the exact same desired properties. In turn, this makes it difficult to compare transformations and measures from one paper to the other. To solve this issue, there is clearly a need for a unified view. 

Another important limitation of the existing work is the lack of control over the original partition and its transformation. Some authors use a single parameter, for example the number of clusters in the transformed partition~\cite{Dom2002, Romano2014}. However, there are other aspects likely to affect the outcome, such as the number and size of the clusters in the original partition, or the intensity of the transformation, and they are not taken into account simultaneously in the literature. This results in a relatively incomplete assessment of the measure properties.

In Section~\ref{subsubsec:CF-applyingTransformation}, we try to solve both these issues, by proposing a unified set of transformations designed to cover most of the literature, and by defining a set of parameters to get the appropriate level of control.

\subsection{Assessment Methods}
\label{subsec:EvaluationApproaches}
After having described the properties that authors want to find in partition comparison measures and the related partition transformations, we now turn to the methods used in the literature to check the presence or absence of these desired properties based on these transformations. We distinguish two families of approaches: visual inspection vs. statistical methods, more specifically correlation and regression.

\subsubsection{Visual Inspection}
\label{subsubsec:VisualInspection}
Visual inspection is perhaps the most intuitive way to characterize the behavior of a measure. Typically, one plots the value of the measure as a function of some parameter used to control the partition transformation, e.g. the number of clusters produced. Authors usually expect a monotonic trend, e.g. proportional increase or decrease in~\cite{Gates2017b}. Some are more specific and look for a specific pattern, e.g. the so-called \textit{knee shape} used in~\cite{Rabbany2013} for a parameter controlling the number of clusters in the transformed partition. It requires the function to reach its maximum when the numbers of clusters in the original and transformed partitions match, and to decrease when there are too few or too many clusters in the transformed partition.

There are mainly two limitations to visual inspection. First, it is not an objective method, so limit cases can be difficult to judge. Second, it can handle only a very limited number of distinct parameters at once, especially if one wants to compare several measures and consider several properties, or assess how parameters interact. Statistical methods allow to solve the first issue, by providing an objective score. There are mainly two types of statistical tools used in the literature to assess measure properties: correlation and regression.

\subsubsection{Correlation}
\label{subsubsec:Correlation}
A \textit{correlation coefficient} quantifies the dependence between two random variables. In our context, and like with visual comparison, these variables are on the one hand the score computed with the measure of interest, and on the other hand a parameter controlling the partition definition or transformation. Many authors~\cite{Horta2015, Zhang2017} use the popular Pearson's product-moment correlation coefficient, which measures the linear dependence between the variables. Others use a rank-order correlation coefficient such as Kendall's (e.g.~\cite{Wu2009}) or Spearman's (e.g.~\cite{Rabbany2013}), which relies on the rank of the values rather than on the values themselves. Compared to Pearson's, such coefficients are able to detect a non-linear dependence, and can thus lead to different conclusions~\cite{Rabbany2013}.

Besides objectivity, another advantage of correlation coefficients over visual inspection is that they summarize the dependence through a single value, which allows representing a number of pairwise relationships in a single table. However, this approach too becomes cumbersome when one wants to consider simultaneously a certain number of parameters and/or measures~\cite{Dom2002}. Moreover, multiple pairwise correlation values are not able to capture the potential interactions between the parameters (i.e. changing one parameter value may affect the partition or transformation feature controlled by another parameter).

\subsubsection{Regression}
\label{subsubsec:Regression}
Regression analysis does not suffer from this limitation, though. In its simplest form, it consists in describing the functional relation between a dependent variable and an independent variable~\cite{Alexopoulos2010}. In our context, those are the considered measure and a parameter of interest, respectively. However, multiple regression allows considering several independent variables at once, i.e. several parameters in our case. Another advantage over correlation is that the regression model can be used not only for interpretation, but also for prediction purposes~\cite{Luo2009}. 

To the best of our knowledge, the work of Saxena \& Navaneetham~\cite{Saxena1991} is the only one that uses multiple regression analysis to assess the similarity of external evaluation measures.
The authors study the effects of three input parameters (cluster size, number of dimensions and number of clusters) on a single measure (the ARI). On top of the regression, they also assess the significance of these effects, and compare the relative importance of the parameters through their associated regression coefficients.

As we will see in Section~\ref{subsubsec:RelativeImportanceAnalysis}, our method goes in the same direction as Saxena \& Navaneetham~\cite{Saxena1991}, but with a more complex model, for the following reasons. First, the set of transformations that we propose in Section~\ref{subsubsec:CF-applyingTransformation} requires to handle more parameters, and therefore to include more independent variables in the model. Second, not only do we study the direct effect of each parameter on the measure, but also their interactions. Third, we consider several distinct measures, and we want to assess and compare the relative importance of the effects that the parameters have on them, which requires a specific processing.


\section{Proposed Framework}
\label{sec:methods}
In this section, we describe the framework that we propose to analyze the behavior of a set of considered measures. It is independent from these measures, so we describe it in a generic way, for any selection of measures.

Our framework is constituted of two parts. The first one consists in characterizing the considered measures through the partition transformation-based principle mentioned in the Introduction (Section~\ref{subsec:evaluationFramework}). The second part is to perform an appropriate regression analysis in order to interpret these characteristics and compare the measures (Section~\ref{subsec:AnalysisFramework}).

\subsection{Characterization of the Measures}
\label{subsec:evaluationFramework}
Our objective is to quantify how similar two partitions are through several external measures, under different scenarios, and then to assess how the resulting values are affected when one of the partitions undergoes systematic and controlled changes. Unlike the common approach taken in the literature, we generate the necessary data in a fully parametric way in order to get a greater control. For the same reason, our approach is deterministic.

\begin{figure}[ht!]
    \captionsetup{width=0.9\textwidth}
	\centering
    \includegraphics[width=0.60\textwidth]{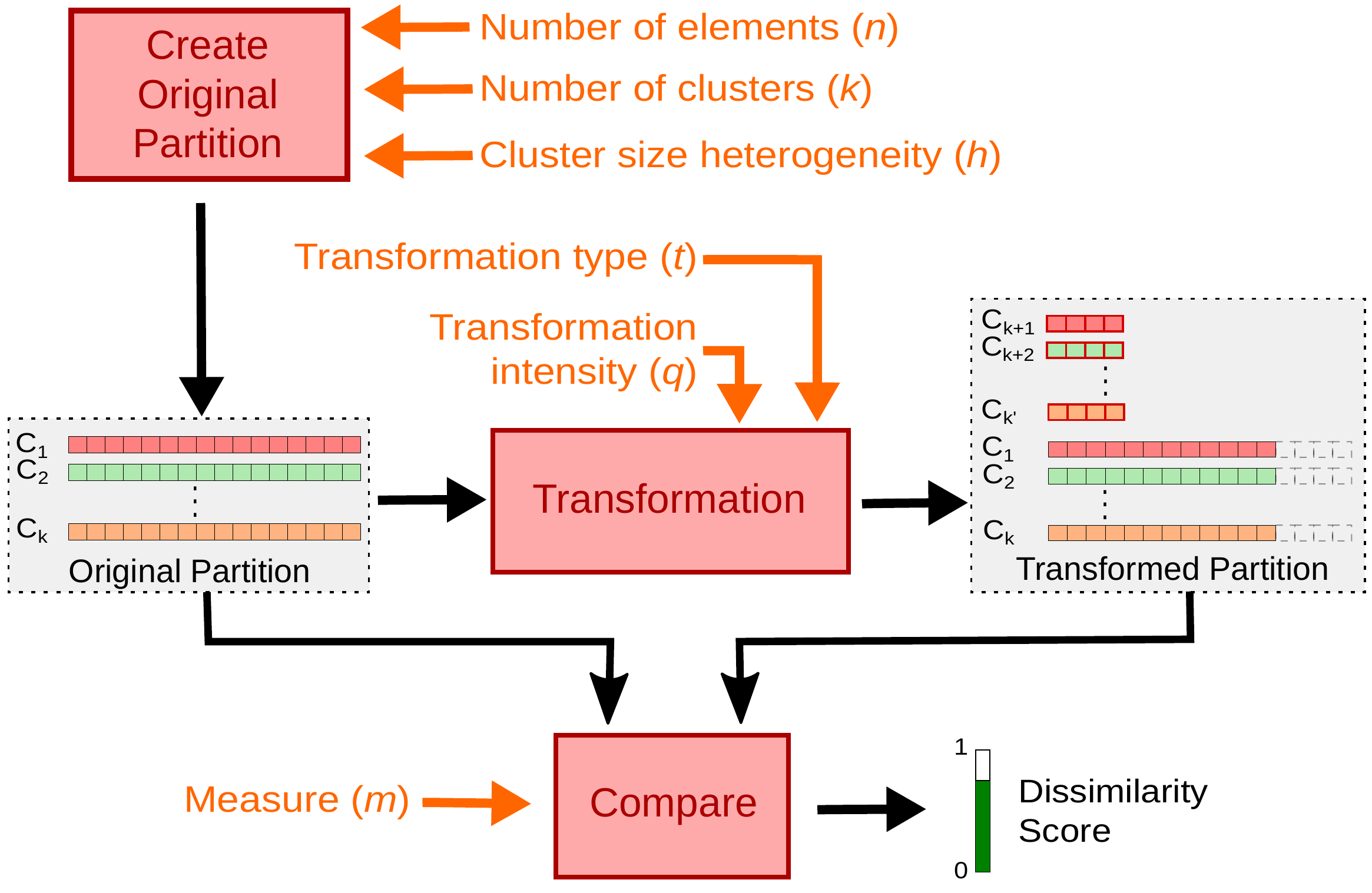}
	\caption{General Framework, with parameters represented in orange. For illustration purposes, the \textit{k New Clusters} transformation is used to produce the output partition with $k$ new clusters. Figure available at \href{https://doi.org/10.6084/m9.figshare.13109813}{10.6084/m9.figshare.13109813} under CC-BY license.}
	\label{fig:general-framework}
\end{figure}

Our three-step method is summarized in Figure~\ref{fig:general-framework}, and
detailed in the rest of this section. The first step is to create a base partition, called \textit{original partition}, and controlled by three parameters (Section \ref{subsubsec:CF-creatingReferencePartition}). The second step consists in applying to this partition a transformation controlled by two other parameters (Section \ref{subsubsec:CF-applyingTransformation}). This leads to a second partition, which we call \textit{transformed partition}. Finally, the third step is to compute the selected external measures in order to assess how similar the original and transformed partitions are (Section \ref{subsubsec:CF-comparisonClusterings}). The whole process is repeated with an adequate number of different parameter values, in order to cover the parameter space.

\subsubsection{Creating the Reference Partition}
\label{subsubsec:CF-creatingReferencePartition}
We control the generation of the reference partition through three parameters: the number of elements $n$, the number of clusters $k$ and the heterogeneity of the cluster sizes $h$. The first two parameters allow to control the most basic aspects of the partition. These are frequently targeted in the literature, albeit not always through explicit parameters. 

The last one is much more uncommon, and lets us control how much cluster sizes vary in the same partition, and therefore to get more realistic cluster sizes. Similar concepts appear in the literature, for example when dealing with balanced vs. imbalanced cluster sizes, but not under the form of such a convenient parameter, to the best of our knowledge. It ranges from $0$ to $1$. When $h=0$, all clusters have the same size (i.e. so-called \textit{balanced} cluster sizes), whereas they get imbalanced when $h>0$, and the differences between their sizes increase when $h$ gets closer to $1$. More formally, the smaller cluster has a size of $s_1 = \alpha$ and the $i$\textsuperscript{th} smallest cluster has a size of $s_i = s_{i-1} + \beta$, whereas $\alpha$ and $\beta$ depend on $k$, $n$ and $h$. In particular, $\beta$ is proportional to $h$. See Appendix~\ref{appendix:DetailsOfClusterSizeHeterogeneity} for details. This choice is a form of compromise allowing to obtain very heterogeneous cluster sizes even for a small $n$ and/or a large $k$.

\subsubsection{Applying the Parametric Transformations}
\label{subsubsec:CF-applyingTransformation}
After having generated the original partition at the previous step, we now want to change it in order to get the transformed partition. Based on our review of the existing work (Section~\ref{subsec:ExistingTransformations}), we propose a set of five parametric transformations aiming at fulfilling several constraints. We want to cover most of the transformations used in the literature, in order to deal with as many desired properties as possible, while keeping our transformations as simple (and thus interpretable) as possible and avoiding overlap between them. We discard \textit{Remove}, as it changes $n$, which in our context is a parameter of the first step of our process. As mentioned before, all our transformations are deterministic in order to offer better control.

We note $t$ the \textit{nature} of the transformation, and use it later as a categorical variable during the regression analysis. We define a parameter $q$ to specify the \textit{intensity} of the transformation, i.e. the proportion of elements it involves. 
It ranges from $0$, meaning no transformation at all, to $1$, in which case the transformation involves all elements.
We want to give the same importance to all clusters when applying the transformation, which means that it affects all of them. However clusters may have different sizes, depending on the heterogeneity of cluster sizes $h$. To deal with this situation, we make the number of elements concerned by the transformation in each cluster proportional to the cluster size.

\begin{figure*}[ht!]
    \captionsetup{width=0.9\textwidth}
	\centering
    \includegraphics[width=0.9\linewidth]{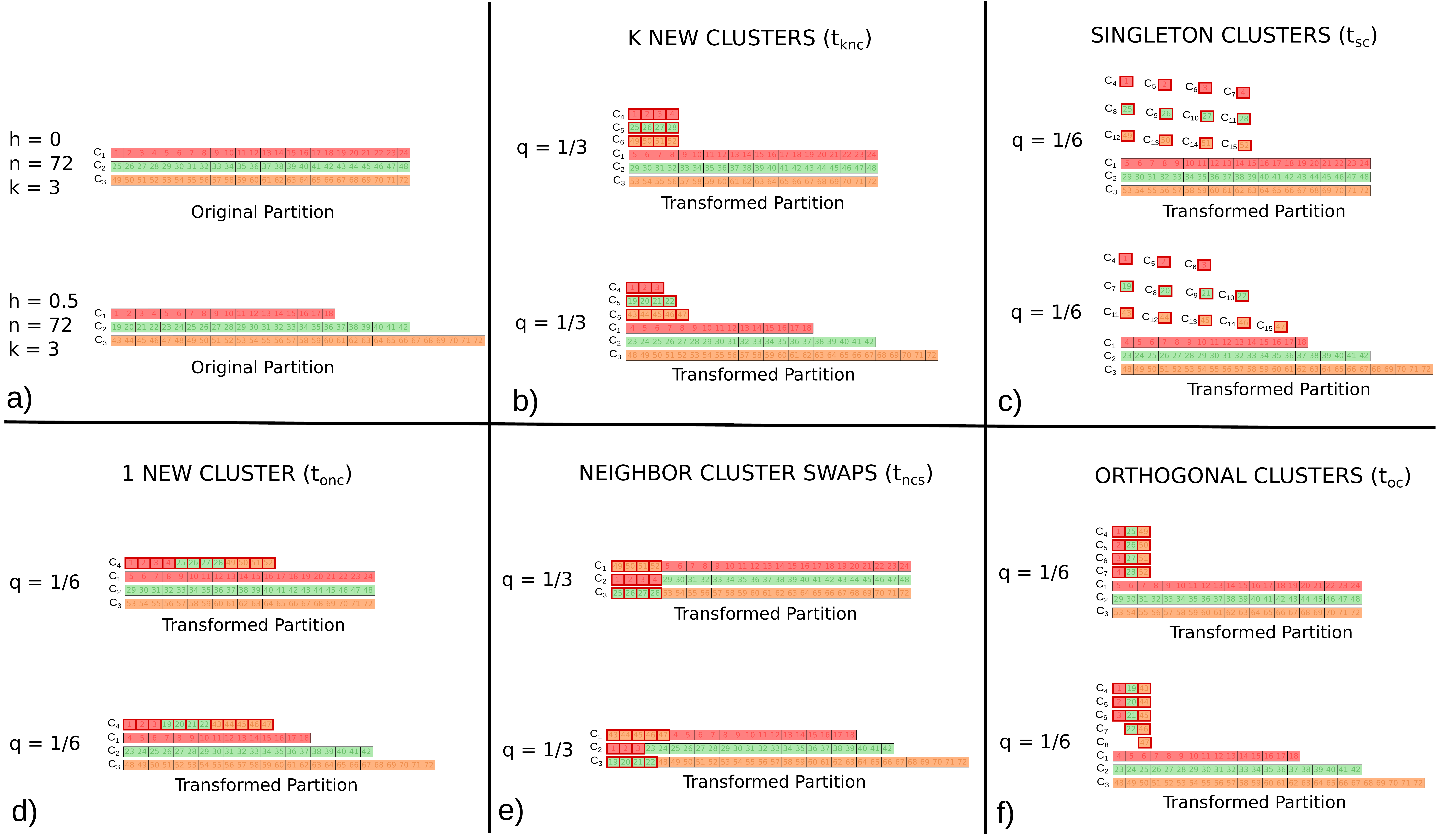}
	\caption{Parametric partition transformations used in our framework. Subfigure a) shows two reference partitions containing both $n=72$ elements and $k=3$ clusters, but differing on the heterogeneity of the cluster sizes: balanced ($h=0$) vs. moderately imbalanced ($h=0.5$). The 5 transformations, illustrated in Subfigures~b)--f), are applied to these two original partitions to produce corresponding transformed partitions. Transformation intensity is $q=1/6$, or equivalently $1/3$ for both transformations concerned with rescaling. Figure available at \href{https://doi.org/10.6084/m9.figshare.13109813}{10.6084/m9.figshare.13109813} under CC-BY license.}
	\label{fig:transformations}
\end{figure*}

The five transformations that we propose are illustrated in Figure~\ref{fig:transformations}, on two example reference partitions (Subfigure~\ref{fig:transformations}a). Both contain $n=72$ elements, represented as numbered squares in the figure, and distributed over $k=3$ clusters, represented by colors. However, the top partition is balanced ($h=0$) whereas the bottom one is moderately imbalanced ($h=0.5$). Each other subfigure shows the partitions resulting from a specific transformation with intensity $q=1/6$. Note that all these transformations allow to test by construction whether or not a measure is sensitive to some framework parameters. On top of that, they can be used to test certain desirable properties from Section~\ref{subsec:DesirableProperties}, as explained in the rest of this section and summarized in Table~\ref{tab:SummaryPropertiesTransformations}.

\paragraph{$k$ New Clusters, $t_{knc}$}
\label{parag:Transf-kNewClusters}
This transformation takes a predefined proportion of each cluster from the original partition, and creates a new cluster with these elements, resulting in $k$ additional clusters (Subfigure~\ref{fig:transformations}b). 
The effect of this proportion on the transformed partition is mirrored in $0.5$. For instance, transforming $40\%$ and $60\%$ of the elements give the same transformed partition. For this reason, we scale $q$ so that it corresponds to twice this proportion, which allows us to keep the same $[0;1]$ range as for the other transformations.

It is worth noting that the transformed partition is a subpartition of the original one, in the sense that each one of its clusters is included in one original cluster. Parameters $k$ and $h$ therefore affect the way the created subclusters relate to the original clusters. This transformation consequently allows testing for the Convex Additivity property, which states that a measure should not be affected when comparing refinements of the same partition. Concretely, we conclude that a measure has this property if it is not affected by $k$ and $h$ when applying this transformation. 

\paragraph{Singleton Clusters, $t_{sc}$}
\label{parag:Transf-SingletonClusters}
All the elements affected by this transformation become singletons, i.e. single-element clusters (Subfigure~\ref{fig:transformations}c). 
This can be viewed as an extreme form of partition refinement, in the sense that each such singleton cluster is fully part of one of the original clusters. Therefore, like $k$ New Clusters, but to a lesser extent, this transformation allows testing the Convex Additivity property through parameters $k$ and $h$. Moreover, it allows checking 
the Sensitivity to Small Differences by considering the effect of parameter $q$. To be consistent with the nature of this property, it is necessary to focus on relatively small values of $q$ (i.e. a limited transformation magnitude), though. 

Parameter $q$ can also be used to assess the Discriminativeness property, as increasing $q$ largely increases the number of clusters in the transformed partition. Therefore, a measure which is affected by an increasing $q$ is likely to discriminate more between transformed partitions whose number of clusters is closer to $k$ (and hence to possess this property~\cite{Rabbany2013}). This is particularly true when the measure scores cover the whole $[0;1]$ range. 
Parameter $k$ can also be used, indirectly, to check the $k$-invariance property. Indeed, the number of clusters created by this transformation does not depend on $k$, and is generally much larger than $k$. So increasing $k$ changes noticeably the number of clusters in the original partition, but not in the transformed one. Consequently, a measure which is marginally or never affected by changes in $k$ when undergoing this transformation can be considered as $k$-invariant. 
%

\paragraph{$1$ New Cluster, $t_{onc}$}
\label{parag:Transf-1NewCluster}
Like the previous transformation, this one takes a proportion of each original cluster, but it gathers these elements to create a \textit{single} cluster instead of $k$ distinct ones (cf. Subfigure~\ref{fig:transformations}d). If we switch the original and transformed partitions, this transformation can alternatively be seen as the removal of a same-sized cluster, i.e. distributing proportionally the elements of a single cluster over the others. This is similar to the transformations used in~\cite{Rezaei2016} 
to test for the Insensitivity to Cluster Size property. In our case, if increasing $k$ results in a substantial change in the measure score (all other things remaining equal), 
then this indicates that the measure is likely to treat the clusters equally, i.e. that it holds the property~\cite{Rezaei2016}.

\paragraph{Neighbor Cluster Swaps, $t_{ncs}$}
\label{parag:Transf-NeighborClusterSwap}
This transformation moves a proportion of each cluster into its neighbor cluster. Each cluster swaps elements with exactly one different cluster (cf. Subfigure~\ref{fig:transformations}e). 
Like for $k$ New Clusters, the effect of this proportion on the transformed partition is mirrored in $0.5$ for certain values of $h$. We therefore rescale it in the same way as before, in order to obtain a parameter $q$ ranging from $0$ to $1$.
This transformation allows to test for the Insensitivity to Cluster Size property through parameter $h$. By design, the number of clusters in the original and transformed partitions are the same. Hence, this transformation does not interfere with $k$ and $h$. If increasing $h$ has a substantial effect on the measure score, then this indicates that the measure is not likely to treat the clusters equally, i.e. it does not hold the property.

\paragraph{Orthogonal Clusters, $t_{oc}$} 
\label{parag:Transf-OrthogonalClusters}
This transformation uses a proportion of each cluster to create new clusters, in such a way that all of their elements come from different original clusters (cf. Subfigure~\ref{fig:transformations}f).  
The resulting clusters are orthogonal to the original ones, in the sense that each original cluster is represented equally in the new clusters.

Applying this transformation with different values of $k$ has an effect on the number of clusters in the transformed partitions, such that the difference in number of clusters between the original and compared partitions substantially decreases, when $k$ increases. This is similar to the transformations used in~\cite{Gates2017b}. 
The main difference is that the authors shuffle completely the transformed partitions, whereas this randomization process is tuned with the parameter $q$ in our case. Therefore, like in~\cite{Gates2017b}, this transformation can be used, to a lesser extent, to test for the $k$-invariance property. If a measure is marginally or never affected by changes in $k$ when undergoing this transformation can be considered as $k$-invariant.

Moreover, this transformation can also be used to check the Proportionality property with parameter $q$, as in~\cite{Liu2019}. If the scores of a distance measure increase linearly with increasing values of $q$, then we say that the measure validates this property. 
Finally, like in \textit{Singleton Clusters}, the Sensitivity to Small Differences property can be also checked through small values of $q$, i.e. a limited transformation magnitude~\cite{Pfitzner2008}.

\begin{table}[h]
\captionsetup{width=0.9\textwidth}
    \footnotesize
	\centering
	\hspace{-0.50cm}
	\footnotesize
    \renewcommand{\arraystretch}{1.5}
    \begin{tabular}{p{2.5cm} | L{2.25cm} | p{9cm}}
    	\hline
    	 Property & Transformation \& Parameter & Description \\
    	 \hline
    	 \multirow{2}{*}{\textit{$k$-invariance}}
    	    & \parbox[l]{2cm}{$t_{sc}$ \& $k$~\cite{Newman2020}} & The measure is marginally affected by changes in $k$ when undergoing this transformation. \\
    	    & \parbox[l]{2cm}{$t_{oc}$ \& $k$~\cite{Gates2017b}} & The measure is marginally affected by changes in $k$ when undergoing this transformation. \\
	     \hline
     	 \multirow{1}{*}{Discriminativeness}
    	    & $t_{sc}$ \& $q$~\cite{Rabbany2013} & Increasing $q$ results in a substantial change in the measure score for this transformation. \\
    	 \hline
     	 \multirow{2}{*}{\parbox[l]{2.5cm}{Insensitivity to Cluster Size}}
    	    & $t_{onc}$ \& $k$ \cite{Rezaei2016} & Increasing $k$ results in a substantial change in the measure score this transformation. \\
    	     & $t_{ncs}$ \& $h$ \cite{Rezaei2016} & The measure is marginally or never affected by this transformation, for increasing $h$. \\
      	 \hline
      	 \multirow{2}{*}{Convex Additivity}
    	    & $t_{sc}$ \& $k$, $h$~\cite{Meila2007} & The measure is not affected by $k$ or $h$ for this transformation. \\
    	    & $t_{knc}$ \& $k$, $h$~\cite{Meila2007} & The measure is not affected by $k$ or $h$ for this transformation. \\
    	\hline
    	\multirow{1}{*}{Proportionality}
    	    & $t_{oc}$ \& $q$ \cite{Liu2019} & The measure score increases proportionally with $q$. \\
    	\hline
    	\multirow{2}{*}{\parbox[l]{2.5cm}{\centering Sensitivity to Small Differences}}
    	    & \parbox[l]{2cm}{$t_{oc}$ \& $q$ \cite{Pfitzner2008}} & Even small values of $q$ have a substantial effect on the measure score. \\
    	    & \parbox[l]{2cm}{$t_{sc}$ \& $q$} & Even small values of $q$ have a substantial effect on the measure score. \\
	    \hline
	\end{tabular}
    \caption{The six desirable properties selected from Section~\ref{subsec:DesirableProperties}, together with the framework parameters and transformations that allow testing them. The bibliographic references indicate matching situations from the literature, when available. 
    }
    \label{tab:SummaryPropertiesTransformations}
\end{table}

\subsubsection{Computing and Normalizing the Measures}
\label{subsubsec:CF-comparisonClusterings}
The third step is very straightforward and simply consists in computing the measures for each pair of partitions generated, in order to compare the reference partition with each transformed partition. Note that during the regression analysis, the measure of interest is considered as a categorical variable noted $m$.

In order for these values to be comparable, one has to make sure they respect two constraints, though. First, some measures of the literature quantify the similarity between two partitions, whereas others assess their \textit{dis}similarity.
For comparison purposes, all measures compared within our framework should express the same concept, be it similarity or dissimilarity. Without loss of generality, we assume in the rest of our framework that all considered measures are dissimilarity measures (possibly after having undergone an appropriate transformation).

Second, all measures are not necessarily defined on the same range, which means that some of them must be normalized in order to allow comparison. Many measures are defined on $[0;1]$, so this seems like a consensual choice.  

\subsection{Regression Analysis}
\label{subsec:AnalysisFramework}

The second part of our framework consists in analyzing all the dissimilarity values obtained during the first part. 
In the following, we first introduce our proposed regression model (Section~\ref{subsubsec:ModelConstruction}). We then turn to relative importance analysis (Section~\ref{subsubsec:RelativeImportanceAnalysis}), which aims at determining how much the framework parameters affect the measures depending on the applied transformations. 

\subsubsection{Model Design}
\label{subsubsec:ModelConstruction}
In our context, the dependent variable is a dissimilarity score in $[0,1]$, which we note $y$, whereas the independent variables correspond to the five parameters of the framework ($n$, $k$, $h$, $q$, $t$) and the nature of the measure used to compute the score ($m$). Four of them are therefore quantitative ($n$, $k$, $h$ and $q$), and two are categorical ($t$ and $m$).

We study the relation between these variables through a multiple linear regression model. Note that in this type of model, the linearity constraint concerns the regression coefficients, and not the independent variables. This means that independent variables can appear as polynomial terms in the model, and that the model can contain interaction terms corresponding to products of independent variables. 
There exist more complex types of regression models (e.g. polynomial regression), which could better fit our data. We chose to use a linear regression nevertheless, because it is much more interpretable~\cite{Cohen2002}, a property which is particularly important in our case.

The presence of \textit{categorical} independent variables makes it necessary to adopt a specific approach, by comparison to a straightforward model including only numeric dependent variables, and there are several methods to do so~\cite{Cohen2002}. 
Among them, we decide to use so-called \textit{dummy variables}, as they allow to avoid splitting the model in several parts, which in turns makes it easier to compare the estimated regression coefficients~\cite{Hardy1993}. 


Our multiple linear regression model is as follows
\begin{equation}
    \footnotesize
    \label{eq:regression-model}
    \begin{split}
    y = \sum_i \sum_j \bigg( & \beta_{0ij} t_i m_j \\
        & + \beta_{1ij} n t_i m_j + \beta_{2ij} k t_i m_j + \beta_{3ij} p t_i m_j + \beta_{4ij} h t_i m_j \\
        & + \beta_{5ij} n k t_i m_j + \beta_{6ij} n h t_i m_j + \beta_{7ij} n p t_i m_j \\
        & + \beta_{8ij} k h t_i m_j + \beta_{9ij} k p t_i m_j + \beta_{10ij} h p t_i m_j \bigg) \\
        & + \epsilon,
    \end{split}
\end{equation}
where the $\beta_{\cdot i j}$ are the regression coefficients, $t_i$ and $m_j$ are the dummy variables, and $\epsilon$ is the common error, which is assumed independent and normally distributed with mean $0$ and standard deviation $\sigma$. Each dummy variable is binary, and represents one specific value of a categorical variable: transformations for $t_i$ ($1 \leq i \leq T$) and measures for $m_j$ ($1 \leq j \leq M$), where $T$ (resp. $M$) is the number of transformations (resp. measures). The model focuses on various types of interactions between the independent variables. The second line contains terms describing interactions between the categorical variables and each \textit{single} numeric variable. The third line deals, in addition, with interactions between \textit{pairs} of quantitative variables. These terms are likely to introduce some amount of collinearity with the corresponding terms from the previous line. In order to solve this issue, we center all the quantitative independent variables~\cite{Kutner2005}. In order to keep the model interpretable, we do not include any higher order term.

\subsubsection{Relative Importance Analysis of Independent Variables}
\label{subsubsec:RelativeImportanceAnalysis}
As this stage, we have a multiple linear regression model able to represent the relations between our framework parameters and the scores of the measures. Next, we want to assess the relative importance (also called relative strength~\cite{Gujarati2003} or effect size~\cite{Trusty2004}) of the terms constituting our model.

In our context, \textit{relative importance} refers to the contribution of an independent variable, by itself and in combination with other independent variables, to the prediction or the explanation of the dependent variable~\cite{Johnson2004}. Such notion can be formalized in a number of ways, therefore several methods have been proposed~\cite{Johnson2004}, originating from different research fields. Nevertheless, they are designed with a common goal in mind, which is to handle both problems frequently occurring in multiple regression analysis and making this task challenging: 1) multi-collinearity between independent continuous variables; and 2) non-linearity of regression models. Since our independent variables are perfectly uncorrelated by design, and since we consider a purely linear model, all of these methods are relatively equivalent in our case. Therefore, we select the most straightforward approach, consisting in using squared standardized regression coefficients (SRC), or \textit{squared $\beta$ weights}~\cite{Johnson2004, Nathans2012}, to assess the relative importance.

When the regression terms are by design perfectly uncorrelated, zero-order correlations and $\beta$ weights are equivalent~\cite{Johnson2004}. Thus, squared $\beta$ weights sum to the explained variance of the dependent variable~\cite{Johnson2004}, generally noted $R^2$. This implies that squared $\beta$ weights can be used as a means of decomposing $R^2$ according to the terms of the model~\cite{Nathans2012}. 
That is, a squared $\beta$ weight close to zero makes a regression term less important, from which we can deduce that it does not play a key role in explaining the observed variance for the dependent variable $y$.

Having a similar beta weight is not sufficient to conclude that two terms have the same importance: the significance of their difference must be statistically tested~\cite{Hardy1993}. In the presence of such significance we can confirm the superiority of the same variable in one transformation type (similarly, for one measure) over the others.
The importance analysis framework includes this test for all pairs of $\beta$ weights.

\section{Experimental Setup}
\label{sec:Experiments}
In order to illustrate how to use our framework and interpret its results, we now apply it to a selection of popular external measures. In this section, we define our experimental setup. We first describe briefly these measures (Section~\ref{subsec:Measures}), before turning to the dataset and the regression assumptions (Section~\ref{subsec:DatasetAndRegressionAssumptions}). The results are presented afterwards, in Section~\ref{sec:Results}.

\subsection{Selected Measures}
\label{subsec:Measures}
In the literature, external measures are divided into three main categories based on the basic principle they rely upon~\cite{Wagner2007, Meila2015}: 1) Pair-counting, 2) Set-matching (or set overlaps) and 3) Information-theory. Among them, the pair-counting measures are the most studied ones. In line with this, for our experimental setup we select 6 widely used measures covering all three categories, with a prevalence of pair-counting measures. The formal description is given in the Appendix (Section~\ref{appendix:EvalMeasures}): in this section, we focus on the principle underlying these measures, as well as their similarities and differences.

A pair of elements can be handled in only two different ways in a given partition: either they belong to the same cluster or to two different clusters. \textit{Pair-counting} measures are based on the idea of comparing how two partitions of the same dataset handle each pair of elements. For a given pair, there is \textit{positive agreement} between the partitions if its elements belong to the same cluster in both partitions; \textit{negative agreement} if they belong to different clusters in both partitions; and \textit{disagreement} otherwise. The \textit{Rand Index} (RI)~\cite{Rand1971} is the proportion of agreement relative to the total number of element pairs. Hubert and Arabie's \textit{Adjusted Rand Index} (ARI)~\cite{Hubert1985} is based on the RI, but additionally includes a \textit{correction for chance}. The \textit{Jaccard Index} (JI) was originally defined to compare sets~\cite{Jaccard1901}, but it is also used as an external measure~\cite{Ben-Hur2001}. It completely ignores negative disagreements, as it corresponds to the proportion of positive agreements relative to the number of disagreements and positive agreements. The \textit{Fowlkes-Mallows Index} (FMI)~\cite{Fowlkes1983} also ignores negative agreements, as it is based on a score corresponding to the proportion of positive agreements relative to the number of pairs belonging to the same cluster \textit{in one partition}. This score is computed separately for each one of the two compared partitions, and the Fowlkes-Mallows Index is the geometric mean of the resulting values.

To represent the category of \textit{set-matching} measures, we select the $F$-measure (F). Note that this name is sometimes used in the literature as a synonym of \textit{harmonic mean}, and therefore covers several distinct measures (e.g.~\cite{Rabbany2013, Gates2017}). We use the definition of Artiles \textit{et al}.~\cite{Artiles2007}, according to which the $F$-measure is the harmonic mean of two quantities called \textit{Purity} and \textit{Inverse Purity}. In order to compute the Purity of a cluster from the first considered partition, one needs first to identify the cluster from the second partition with which it has the largest intersection. The Purity then corresponds to the proportion of the first cluster which belongs to this intersection. The Purity of the first partition is the total purity of its clusters. The Inverse Purity is simply the Purity of the second partition relative to the first. Finally, the $F$-measure is the harmonic mean of the Purity and Inverse Purity.

\textit{Information-theoretical} measures are generally based on the notion of \textit{Mutual Information}~\cite{Cover2006}. The principle behind these measures is to consider each partition as a categorical random variable, whose possible values are the clusters. The mutual dependence between these variables can then be interpreted as the similarity between the partitions. There are a number of variants of the notion of mutual information, in particular several normalizations have been proposed (see for instance~\cite{Vinh2010}). In this work, we focus on the sum normalization 
as defined in~\cite{Strehl2002}, which is very widespread, and results in the so-called \textit{Normalized Mutual Information} (NMI).

As mentioned in Section~\ref{subsubsec:CF-comparisonClusterings}, our framework expects that all measures express the same concept, either dissimilarity or similarity, and that they are all defined on the same fixed range. Regarding the latter point, all the selected measures are originally ranging from $0$ to $1$ except ARI, which can output negative values in theory. However, in practice it is very rare to get negative values for ARI. In the context of our experiments, it is always positive, so we decided not to perform any additional change. Regarding the former point, we adjust our selected measures through a simple subtraction, so that they all quantify the \textit{dissimilarity} between partitions. We note the resulting measures as follows: $D_{RI}$ (Rand Index), $D_{ARI}$ (Adjusted Rand Index), $D_{FMI}$ (Fowlkes-Mallows Index), $D_{JI}$ (Jaccard Index), ${D_F}$ ($F$-measure) and $D_{NMI}$ (Normalized Mutual Information).

\subsection{Dataset and regression assumptions}
\label{subsec:DatasetAndRegressionAssumptions}
We generate our data through the process presented in Section~\ref{subsec:evaluationFramework}, using the following parameter values. For the number of elements $n$, we choose values arithmetically compatible with the desired numbers of clusters, ranging from $3,240$ 
to $12,960$ with increments of $1,080$. The number of clusters $k$ ranges from $2$ to $11$. The heterogeneity of the clusters size $h$ ranges from $0$ to $0.9$ with increments of $0.1$. Regarding the transformations, their intensity $q$ ranges from $0.1$ to $1$, also by increments of $0.1$, and the nature $t$ of the transformation itself is one among $t_{sc}$ (\textit{Singleton Clusters}), $t_{onc}$ (\textit{1 New Cluster}), $t_{knc}$ (\textit{$k$ New Clusters}), $t_{ncs}$ (\textit{Neighbor Cluster Swaps}), $t_{oc}$ (\textit{Orthogonal Clusters}), as defined in Section~\ref{subsubsec:CF-applyingTransformation}. In the end, the different combinations of our parameter values produce a total of $50,000$ pairs of partitions. 

There are several standard assumptions to check before performing a linear regression~\cite{Gujarati2003, Kutner2005, Cohen2002}: 1) sufficient sample size, 2) linear relationships, 3) no or little multicollinearity, 4) multivariate normality, and 5) homoscedasticity. We review them here for our dataset and framework.
First, our sample size of $50,000$ observations is large enough for getting reliable estimates of the regression.
Second, after a visual inspection we observe that the relation between the dependent variable and the independent variables appear to be linear, except for $k$ and $q$ in which case it looks rather curvilinear. We stick to the linear model for the sake of readability and understandability, though.
Third, by design of our dataset, the observations are independent and there is no collinearity between the independent variables.
Fourth, the large size of our dataset makes the possible presence of outliers unlikely to affect our results~\cite{Faraway2002}.
For the same reason, the central limit theorem guarantees that the residuals will be approximately normally distributed.
Fifth, a visual inspection reveals that the variance of $y$ increases with parameters $q$ (intensity of the transformation) and $k$ (number of clusters), which means the data are not completely homoscedastic. The standard way of solving this issue is to introduce non-linear terms in the model~\cite{Gujarati2003, Kutner2005, Cohen2002}, but again we want to keep it simple, and moreover the observed level of heteroscedasticity does not prevent us from interpreting the regression coefficients~\cite{Gujarati2003}.

\section{Results and Discussion}
\label{sec:Results}
We now assess, compare and discuss the performance of the considered measures when applied to the generated dataset. We first show the relevance of our method through visual inspection (Section~\ref{subsec:VisualInspection}), then present our results in further detail (Section~\ref{subsec:ResultsRelativeImportance}).

\subsection{Visual inspection}
\label{subsec:VisualInspection}
\begin{figure*}[ht!]
\captionsetup{width=0.9\textwidth}
\centering
    \begin{subfigure}[b]{0.3\linewidth}
        \includegraphics[width=\linewidth]{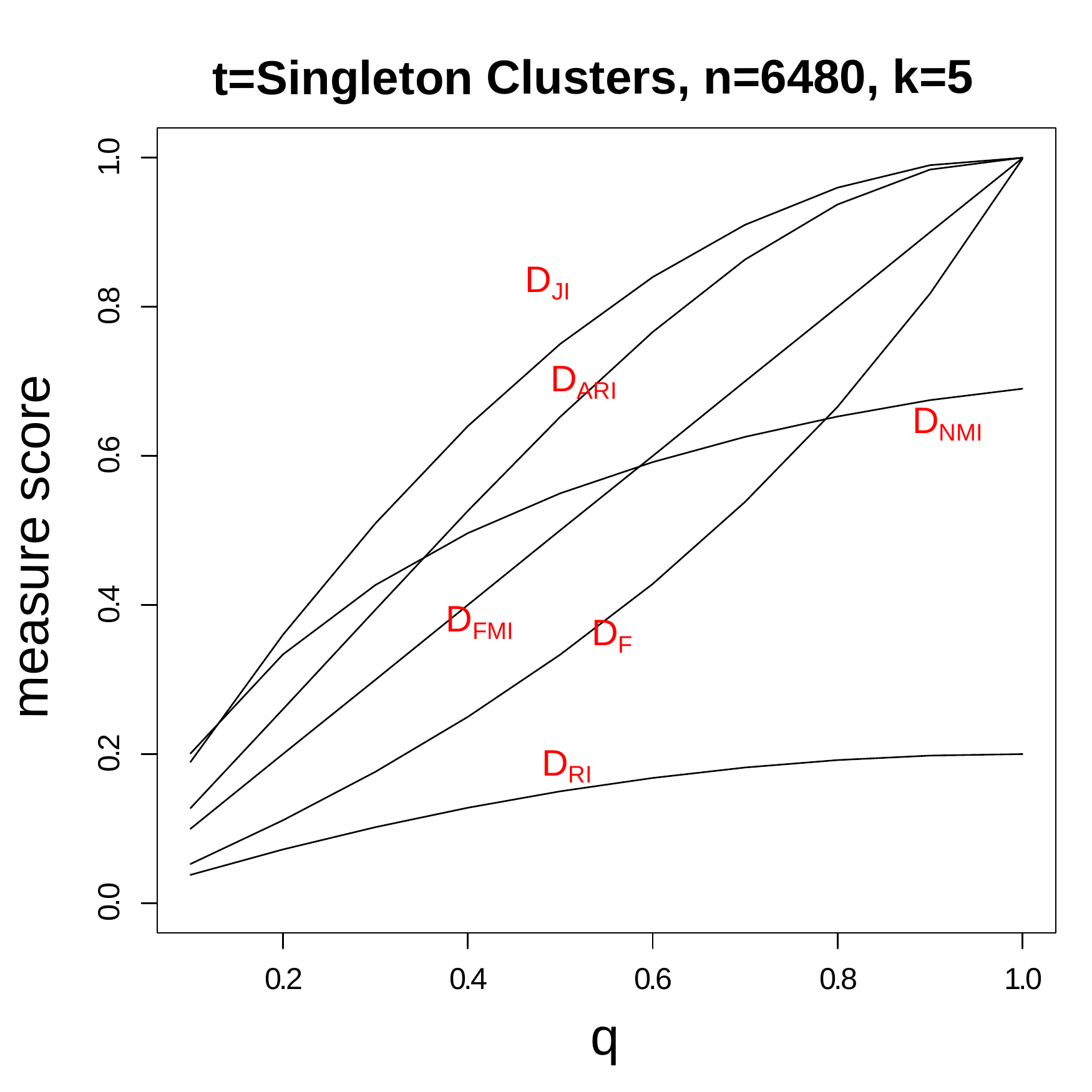}
        \caption{}
        \label{subfig:VisualResult-SlopCoefficients_with_q}
    \end{subfigure}
    \begin{subfigure}[b]{0.3\linewidth}
        \includegraphics[width=\linewidth]{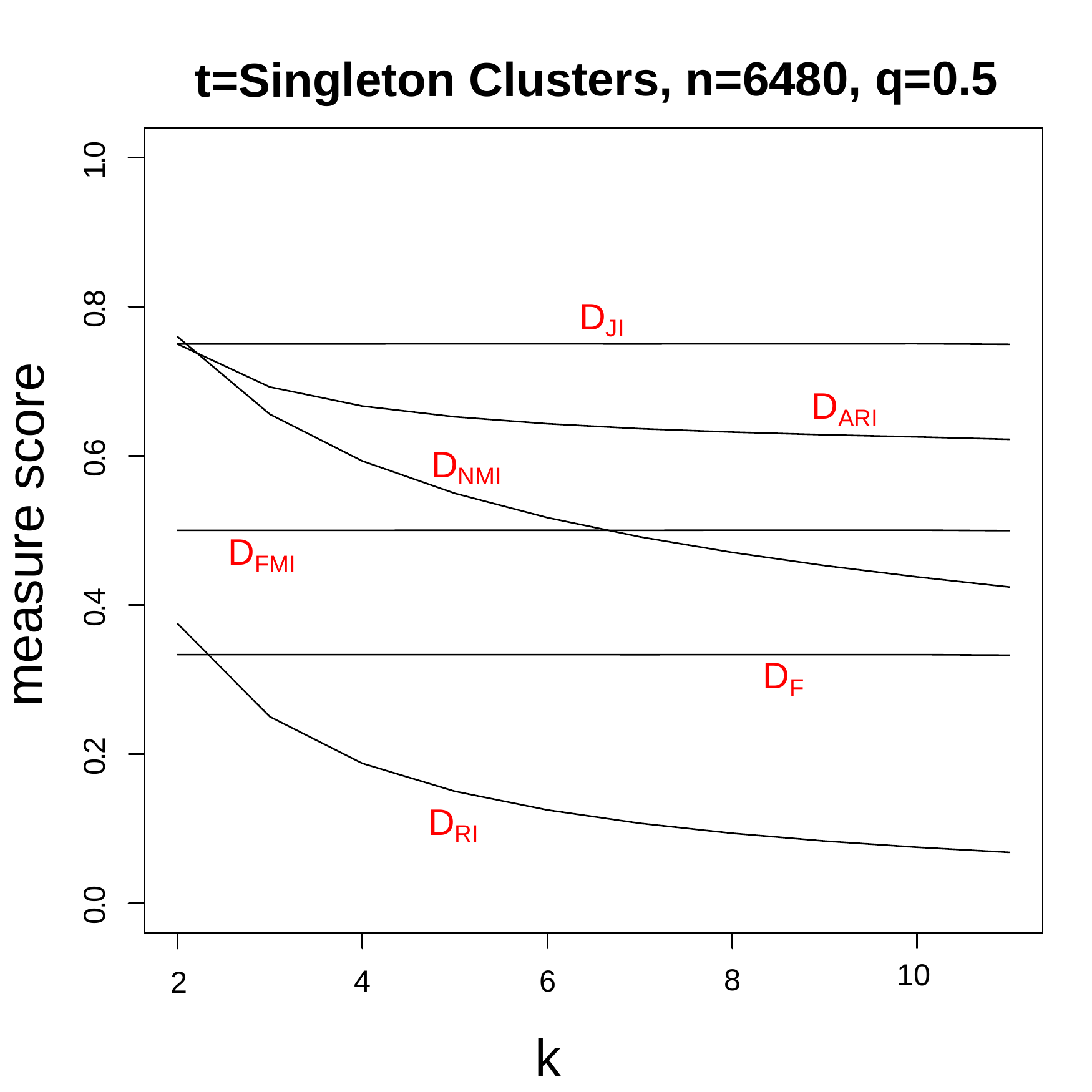}
        \caption{}
        \label{subfig:VisualResult-SlopCoefficients_with_k}
    \end{subfigure}
    \begin{subfigure}[b]{0.3\linewidth}
        \includegraphics[width=\linewidth]{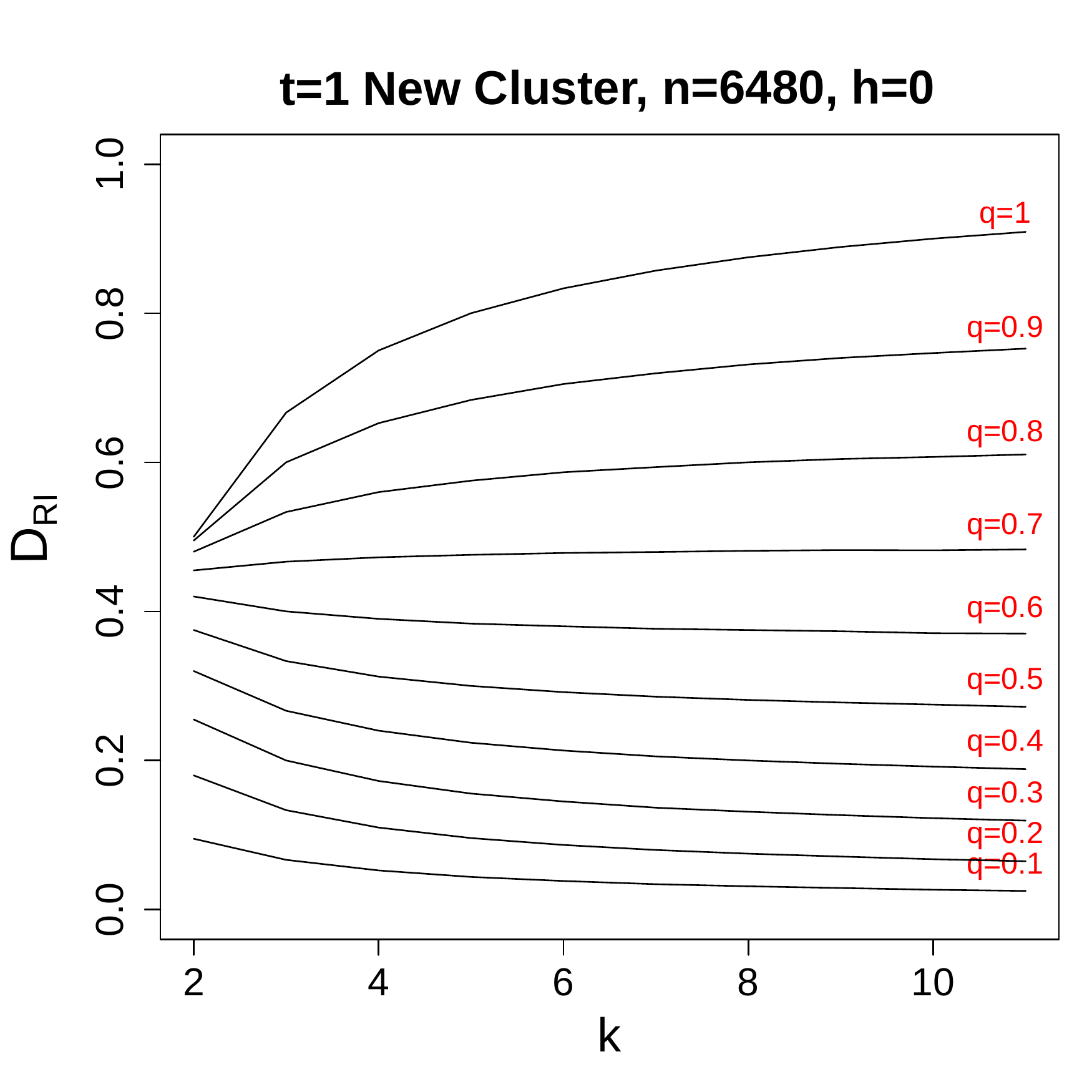}
        \caption{}
        \label{subfig:VisualResult-NonMonotoneTrend}
    \end{subfigure}
	\caption[Fig]{(a) Score of each measure as a function of $p$, for the \textit{Singleton Clusters} transformation. (b) Score of each measure as a function of $k$, for the \textit{Singleton Clusters} transformation. (c) $D_{RI}$ score as a function of $k$, for the \textit{1 New Cluster} transformation, and for several values of $q$. Figures available at \href{https://doi.org/10.6084/m9.figshare.13109813}{10.6084/m9.figshare.13109813} under CC-BY license.}
	\label{fig:VisualResults}
\end{figure*}

To show the relevance of our method we highlight two aspects of our analysis through the visual inspection of Figure~\ref{fig:VisualResults}: 1) slope coefficients and 2) interaction between parameters. As we will see in Section~\ref{subsec:ResultsRelativeImportance}, those aspects allow our method to identify similarities and differences between the considered measures, and therefore to discriminate between them.


Plot~\ref{subfig:VisualResult-SlopCoefficients_with_q} shows how the measure scores evolve as functions of $q$, for the \textit{Singleton Clusters} transformation, while the other parameters are fixed to arbitrary values. One can observe that all the scores increase with $q$, albeit in different ways. Overall, $D_{RI}$ has the smallest slope coefficient, followed by $D_{NMI}$, and they are therefore the least sensitive to this transformation. We observe that $D_{JI}$, $D_{ARI}$, $D_{FMI}$ and $D_F$ get similar scores for extreme $q$ values, but are relatively different when $q$ gets closer to $0.5$. Plot~\ref{subfig:VisualResult-SlopCoefficients_with_k} is built upon the same principle, except it focuses on $k$ instead of $q$. As before, all measures differ in terms of absolute score values, but this time one can detect similar certain trends. In particular, $D_{JI}$, $D_{FMI}$ and $D_F$ remain unchanged, whereas $D_{RI}$, $D_{ARI}$ and $D_{NMI}$ decrease with $k$. These two plots show that our framework is able to produce situations for which the measures behave differently. Moreover, they also show that the slope coefficients, which constitute the basis of our analysis, are able of capturing these differences.

As mentioned in Section~\ref{subsubsec:VisualInspection}, the common way to assess the performances of the measures is trough visual inspection, which requires fixing many parameters, as we did just now, as such plots are able to handle only a limited number of parameters at once. Plot~\ref{subfig:VisualResult-NonMonotoneTrend} illustrates the limitation of this approach by showing the evolution of the $D_{RI}$ score for the \textit{1 New Cluster} transformation, as a function of both $k$ and $q$. When considering only $k$, the $D_{RI}$ score is always monotonic. However the nature and slope of the trend depend on $q$: increasing for $q \geq 0.7$ vs. decreasing for $q < 0.7$. This means that there is an interaction between both parameters. This type of joint effect between parameters is hard to detect when using only plots, as it requires considering all possible combinations of parameters. However, it is captured by the interaction terms present in our regression model, as we will see in Section~\ref{subsec:ResultsRelativeImportance}.

\subsection{Relative importance analysis}
\label{subsec:ResultsRelativeImportance}
We first discuss the effect of the framework parameters on each measure (Section~\ref{sec:EffectParams}), and compare them. Along with our discussion, we identify the desirable properties possessed by each measure, as well. We then show how this analysis can be leveraged to derive a typology of the measures (Section~\ref{sec:TypoMeasures}).

\subsubsection{Effect of the Parameters}
\label{sec:EffectParams}
\begin{figure*}[ht!]
    \captionsetup{width=0.9\textwidth}
	\centering
    \includegraphics[width=0.9\linewidth]{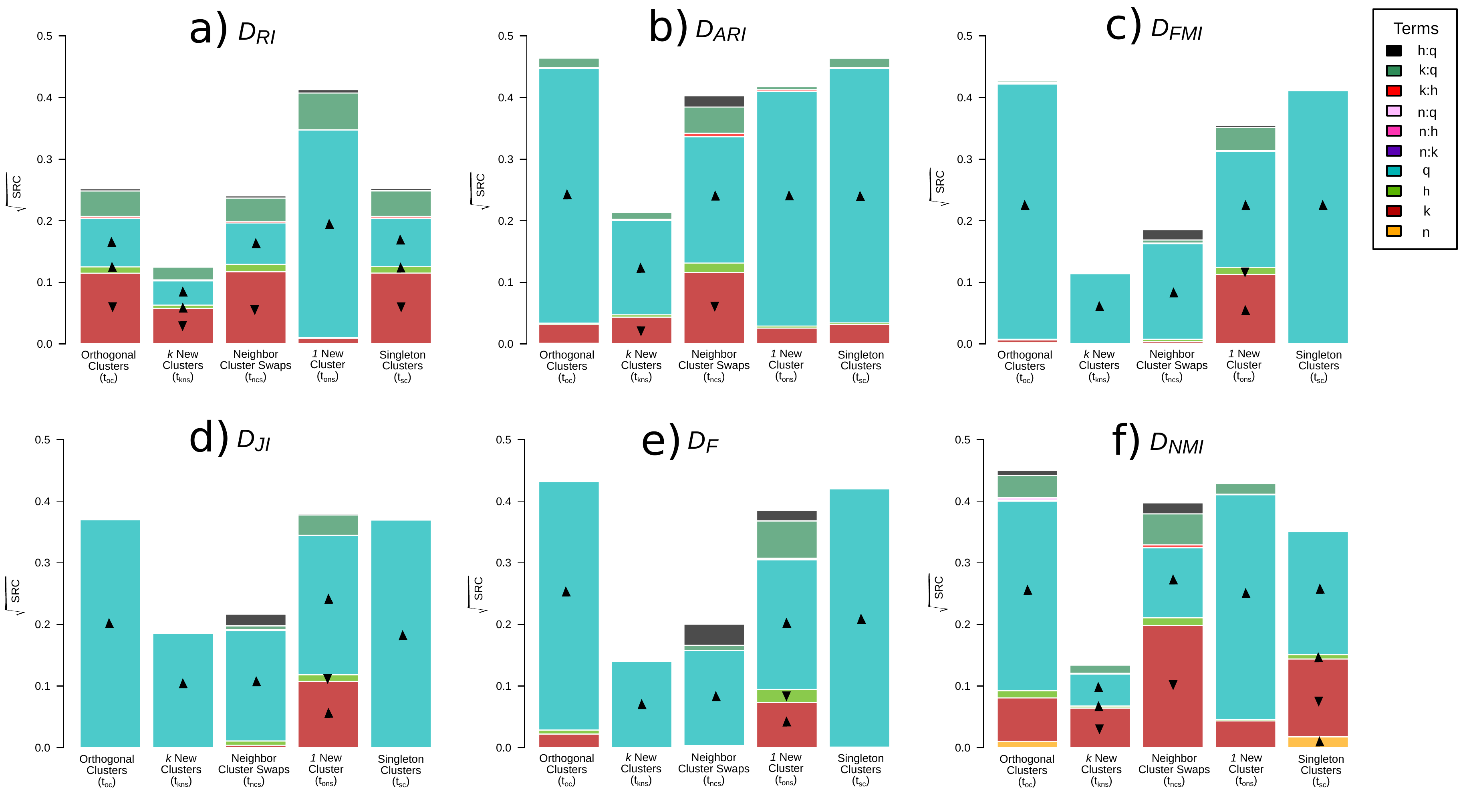}
	\caption{Results of the relative importance analysis, for measures a) $D_{RI}$, b) $D_{ARI}$, c) $D_{FMI}$, d) $D_{JI}$, e) $D_{F}$ and f) $D_{NMI}$. The order of the terms in each bar is shown in the legend. The relative importance scores represented on the $y$-axis are square-roots. Upper (resp. lower) triangles indicate an increasing (resp. decreasing) trend of measure scores, when the corresponding parameter increases, independently of the values of the other parameters. Figure available at \href{https://doi.org/10.6084/m9.figshare.13109813}{10.6084/m9.figshare.13109813} under CC-BY license.}
	\label{fig:RelativeImportanceResults}
\end{figure*}

We show all the results from our relative importance analysis in Figure~\ref{fig:RelativeImportanceResults}, using stacked barplots. We describe these plots globally here, for matters of convenience, before interpreting them in the rest of this section. The figure contains 6 barplots (i.e. subfigures), each one corresponding to a specific dissimilarity measure. Each barplot is constituted of 5 stacked bars, each one corresponding to a different transformation. The segments constituting these stacked bars represent the regression terms from \eqref{eq:regression-model}.
Their colors and order match the legend, and their height corresponds to the associated regression coefficient $\beta$ in \eqref{eq:regression-model}.
More precisely, the segment heights are proportional to the square root of the squared $\beta$ coefficients.

The larger the segment height, the more important the regression term for the measure and transformation represented by the considered stacked bar. The values they represent are unitless, and we perform no upper bound normalization in order to ease comparisons between transformations and measures. Differences between segment heights are not always statistically significant, though. The exhaustive list of significant differences at $p$-value $\leq 0.05$
is given in Appendix (Figures~\ref{fig:Appendix_significance_diffrences_for_measures} and~\ref{fig:Appendix_significance_diffrences_for_transfs}) for the sake of completeness. However, we find it difficult for the reader to cross-check them systematically with Figure~\ref{fig:RelativeImportanceResults}. It is more intuitive to use the following rule of thumb: if one can visually detect a difference between two bars of Figure~\ref{fig:RelativeImportanceResults}, then it is statistically significant. 

Finally, there is a last bit of information in Figure~\ref{fig:RelativeImportanceResults}, under the form of triangles placed over certain segments and representing monotonic behaviors. Upward (resp. downward) triangles indicate that the measure score consistently increases (resp. decreases) when the concerned parameter increases, independently from the other parameters. This information can be seen as complementary to the relative importance analysis. Suppose that a given parameter is similarly important for several measures, i.e. it affects them to roughly the same extend. The triangles allow distinguishing the measures qualitatively, based on the nature of this effect (see Section~\ref{sec:PracticalCases} for a practical example).

Overall, we can observe that all measures are strongly affected by $q$, and to a lesser extent by $k$ and $h$. On the contrary, $n$ has close to no effect on the measures. This effect of $q$ on all measures also appears under a different form in Figure~\ref{subfig:VisualResult-SlopCoefficients_with_q}. As shown by the triangles in Figure~\ref{fig:RelativeImportanceResults}, the measure score increases with $q$ in all cases. This general behavior is intuitively sound, as $q$ controls the intensity of the transformation. There are differences, as illustrated in Figure~\ref{subfig:VisualResult-SlopCoefficients_with_q}, in the way the measures are affected by $q$ and the other parameters, though, and we can also see some punctual effects due to interactions between parameters. 
In the following, we consider each measure and discuss the results displayed in Figure~\ref{fig:RelativeImportanceResults}.

\paragraph{RI}
We can distinguish roughly two categories of transformations regarding $D_{RI}$, depending on how the measure is affected by the parameters. The first category contains only \textit{1 New Cluster}, for which we observe a sensitivity almost twice as important as for the other transformations, which is unique among the considered measures. Also, this transformation exhibits a very strong effect of $q$, and to a lesser extent of the interaction between $q$ and $k$, as also illustrated in Figure~\ref{subfig:VisualResult-NonMonotoneTrend} from a different perspective. The second category contains the rest of the transformations, for which parameter importance is more balanced between $q$, $k$, and their interaction. 

As mentioned in Section~\ref{subsec:Measures}, $D_{RI}$ considers positive and negative agreements equally. When applying a transformation of the second category, an increase in $q$ causes the number of positive agreements to decreases, whereas the negative agreements are largely preserved. This prevents $D_{RI}$ from using its whole nominal range $[0, 1]$, as also pointed out by Meil{\u{a}}~\cite{Meila2015} and Vinh \textit{et al}.~\cite{Vinh2010}. 
This in turns explains the observed smaller effect of $q$. As explained in Section~\ref{parag:Transf-SingletonClusters}, we can infer from a small $q$ effect for \textit{Singleton Clusters} that $D_{RI}$ does not possess the Discriminativeness property, a conclusion that confirms the results of Rabbany \textit{et al}.~\cite{Rabbany2013}.

On the contrary, there is a relatively substantial effect of $k$ for the transformations of the second category, which is due to them largely preserving negative agreements, as already noticed for $q$. As explained in Section~\ref{subsubsec:CF-applyingTransformation}, the large effect of $k$ for the \textit{Singleton Clusters} and \textit{Orthogonal Clusters} transformations indicates that the measure is not $k$-invariant. Similarly, the large effect of $k$ for the \textit{Singleton Clusters} and \textit{$k$ New Clusters} transformations indicates that it does not have the Convex Additivity property. These findings are in line with the results of Rabbany \textit{et al}.~\cite{Rabbany2013} and Amelio \& Pizzuti~\cite{Amelio2016} regarding $k$-invariance, and Meil{\u{a}}~\cite{Meila2015} regarding Convex Additivity.

The relatively small effect of $k$ for \textit{1 New Cluster} shows that $D_{RI}$ is sensitive to variations in the cluster sizes (cf. no Insensitivity to Cluster Size), as already pointed out by Rezaei \& Fränti~\cite{Rezaei2016} for pair-counting measures. 
The absence of any significant effect of $h$ in the results of \textit{Neighbor Cluster Swaps} corroborates this finding.
Regarding the remaining parameters, \textit{1 New Cluster} is also the only transformation which seems not to be affected by $h$. Finally, $n$ does not seem to affect $D_{RI}$ at all.

\paragraph{ARI}
Overall, $q$ has a much stronger effect on $D_{ARI}$ when compared to $D_{RI}$, which results in a total sensitivity approximately twice as large for all transformations except \textit{1 New Cluster} (which is already large in $D_{RI}$). Based on the effects observed for \textit{Singleton Clusters}, $D_{ARI}$ seems to validate the Discriminativeness property much more than $D_{RI}$.

The effect of $k$ is much lower than in $D_{RI}$, for all transformations except \textit{Neighbor Cluster Swaps}. According to certain results obtained by Meil{\u{a}}~\cite{Meila2015} for a similar transformation, the correction term in $D_{ARI}$ can be sensitive to variations in the cluster number and sizes, which may explain our observation. 
Regarding \textit{1 New Cluster}, the effect of $k$ in $D_{RI}$ is already small, and the correction present in $D_{ARI}$ only slightly increases it. Therefore, $D_{ARI}$ is sensitive to the variations of the cluster sizes, too (cf. Section~\ref{parag:Transf-1NewCluster}). 

The effect of $k$ observed for \textit{Singleton Clusters} and \textit{Orthogonal Clusters} is much smaller than in $D_{RI}$, which indicates that the measure is $k$-invariant (Sections~\ref{parag:Transf-SingletonClusters} and \ref{parag:Transf-OrthogonalClusters}). This is consistent with the fact that $D_{ARI}$ was designed specifically to make $D_{RI}$ $k$-invariant, a property already verified empirically by Rabbany \textit{et al}.~\cite{Rabbany2013}. However, this effect is still noticeable, which shows that the measure is not completely $k$-independent. Similarly, the effect of $k$ for \textit{$k$ New Clusters} is smaller than in $D_{RI}$ but still considerable. Based on these two observations, we can conclude that $D_{ARI}$ does not possess the Convex Additivity property (Section~\ref{parag:Transf-kNewClusters}).

The introduction of chance correction has a side-effect on $h$, as it has a much smaller effect on $D_{ARI}$ compared to $D_{RI}$, for all transformations. 
This is consistent with a similar observation pointed out by Romano \textit{et al}.~\cite{Romano2016}. 
Interaction-wise, the effect of $k$:$q$ is much weaker than in $D_{RI}$, probably due to the lower overall effect of $k$, except for \textit{1 New Cluster}. Finally, $n$ does not seem to affect $D_{ARI}$ at all.

\paragraph{FMI \& JI}
We jointly discuss both other pair-counting measures, $D_{FMI}$ and $D_{JI}$, because their results are very similar and differ only on the magnitude of the effect of $q$. The main difference with the other measures is that $q$ is the only perceptible effect for three transformations: \textit{Orthogonal Clusters}, \textit{k New clusters} and \textit{Singleton clusters}. Consequently, both measures differ from the two previous ones regarding certain desirable properties. First, like $D_{ARI}$ but unlike $D_{RI}$, both measures possess the Discriminativeness property. Second, unlike $D_{RI}$ and $D_{ARI}$, they seem to validate the Convex Additivity property. It is worth stressing that, in theory, $D_{FMI}$ and $D_{JI}$ are not supposed to possess this last property~\cite{Meila2007}, \textit{strictly} speaking.  
However, our results show that in practice they behave as if they do, at least \textit{to some extent}, and under some conditions (here: when the number of elements $n$ is large enough).

The effect of $k$ is negligible for all transformations but \textit{1 New Cluster}, i.e. the second category of transformations previously identified for $D_{RI}$. These transformations affect only marginally negative agreement, which explains why the effect of $k$ is so small here, compared to $D_{RI}$. This effect is small for \textit{Singleton Clusters} and \textit{Orthogonal Clusters}, so we can conclude that both measures appear to validate the \textit{$k$-invariance property} (Sections~\ref{parag:Transf-SingletonClusters} and \ref{parag:Transf-OrthogonalClusters}). The strong effect of $k$ for \textit{1 New Cluster} indicates that these measures possess the Insensitivity to Cluster Size property (Section~\ref{parag:Transf-1NewCluster}). 

Regarding the other effects, one can observe that unlike $D_{RI}$ and $D_{ARI}$, $h$ has a small effect only for \textit{1 New Cluster}. Furthermore, not only do $k$ and $q$ have a strong effect for this transformation, but their interaction does too. Finally, overall, $n$ has no significant effect on both measures.

\paragraph{$F$-measure}
Unlike the previous measures, which rely on pair-counting, $D_{F}$ is based on set-matching. Nevertheless, the observed effects are very similar to those of $D_{FMI}$ and $D_{JI}$. We observe essentially two differences. The first is that $k$ and $h$ have a relatively noticeable effect for \textit{Orthogonal Clusters}. The second is that the effect of interaction $h$:$q$ is stronger for \textit{Neighbor Cluster Swaps} and \textit{1 New Cluster}. $D_{F}$ still validates the same properties as $D_{FMI}$ and $D_{JI}$ do, despite these small differences.

\paragraph{NMI}
The results obtained for the information-theoretical measure $D_{NMI}$ are very similar to those of $D_{RI}$, qualitatively speaking, and to those of $D_{ARI}$, in terms of magnitude of the effect observed for each transformation. Like $D_{RI}$, $D_{NMI}$ behaves in the same way for all the four desirable properties,
and this is consistent with the observations from the literature. For instance, Meil{\u{a}}~\cite{Meila2007} proves that the rescaling performed by some measures for normalization purposes, such as $NMI$, have the effect of breaking the Convex Additivity property. Moreover, Newman~\textit{et al.}~\cite{Newman2020}, like others~\cite{Vinh2010, Rabbany2013, Amelio2016}, show that $NMI$ tend to favor partitions with more clusters when compared with a reference partition (cf. \textit{no} $k$-invariance), and that this behavior can be smoothed by correcting \textit{NMI} for chance. 


A clear difference between $D_{NMI}$ and all the other measures is that $n$ has a very visible effect for \textit{Orthogonal Clusters} and \textit{Singleton Clusters}. This seems to be an artefact of the normalization for these transformations, which would match the observation made by Amelio \& Pizzuti~\cite{Amelio2016}, 
rather than a violation of the $n$-invariance property. Indeed, the information-theoretic measures are $n$-invariant by construction~\cite{Meila2007}.

\paragraph{General Observations}
For the sake of clarity, we roughly summarize in Table~\ref{tab:SummaryPropertiesMeasures} the discussion that takes place throughout the current section regarding the presence or absence of desirable properties within the considered measures. We observe that three measures validate all 4 properties ($D_{F}$, $D_{JI}$, $D_{FMI}$), whereas two measures have none of them ($D_{RI}$, $D_{NMI}$). The last one, $D_{ARI}$, holds an intermediary position, as it possesses the $k$-invariance and Discriminativeness properties like $D_{F}$, $D_{JI}$ and $D_{FMI}$, whereas it shares the same behavior with $D_{RI}$ and $D_{NMI}$ regarding Insensitivity to Cluster Size and Convex Additivity.

Let us now conclude this section by highlighting the main observations we could draw from the relative importance analysis. First, it is important to stress that the results produced by our framework are consistent with those published in the literature, including both theoretical and empirical works. This is summarized in Table~\ref{tab:SummaryPropertiesMeasures}. Second, the systematic nature of our approach helps uncovering properties not already described in the literature. For instance, Rezaei \& Fränti~\cite{Rezaei2016} state that set matching measures are more suitable regarding the Insensitivity to Cluster Size property. Nevertheless, we find out that the pair-counting measures $D_{JI}$ and $D_{FMI}$ also possess this property. Third, our framework allows us to state that some measures possess certain properties at least partially, or under certain conditions. Indeed, our framework does not predict the presence of a property in a Boolean way, but rather on some continuous spectrum, through regression. Put differently, instead of predicting whether a measure has a property or not, we can estimate \textit{how much} it possesses this property, and assess how this can change depending on the parameter values. For instance, as mentioned above, we can say that $D_{ARI}$ validates the Discriminativeness property much more than $D_{RI}$, based on the effect of $q$ for \textit{Singleton Clusters}. 

\begin{table}[h]
\captionsetup{width=0.9\textwidth}
    \footnotesize
	\centering
	\begin{tabular}{l l l l l}
    	\hline
    	  & $k$-invariance & Discriminativeness & Insensitivity to Cluster Size & Convex Additivity \\
    	  & ($t_{sc}$ and $t_{oc}$ with $k$) 
    	    & ($t_{sc}$ with $q$) 
    	        & ($t_{onc}$ with $k$, $t_{ncs}$ with $h$) 
    	            & ($t_{sc}$ and $t_{knc}$ with $k$ and $h$) \\
    	 \hline
    	 $D_{RI}$ & $\xmark$~\cite{Rabbany2013, Amelio2016, Meila2015}  & $\xmark$~\cite{Rabbany2013} & $\xmark$~\cite{Souto2012, Rezaei2016} & $\xmark$~\cite{Meila2015} 
    	    \\
     	 $D_{ARI}$ & $\cmark$~\cite{Rabbany2013} & $\cmark$~\cite{Rabbany2013} & $\xmark$~\cite{Souto2012, Rezaei2016} & $\xmark$~\cite{Meila2015} \\
    	 $D_{FMI}$ & $\cmark$~\cite{Gates2017b} & $\cmark$ & $\cmark$ & $\cmark$ \\
         $D_{JI}$ & $\cmark$~\cite{Gates2017b} & $\cmark$~\cite{Rabbany2013} & $\cmark$ & $\cmark$ \\
         $D_{F}$ & $\cmark$ & $\cmark$ & $\cmark$~\cite{Souto2012} 
                                                                & $\cmark$ \\
         $D_{NMI}$ & $\xmark$~\cite{Vinh2010, Rabbany2013, Amelio2016, Gates2017b, Newman2020} & $\xmark$~\cite{Rabbany2013, Amelio2016} & $\xmark$~\cite{Souto2012, Rezaei2016} & $\xmark$~\cite{Meila2015} \\
    	\hline
	\end{tabular}
    \caption{Relations between four desirable properties and the considered measures, based on our results presented in Figure~\ref{fig:RelativeImportanceResults}. The method used to check whether a measure has a property is summarized between parenthesis in the first line, and additional details can be found in Section~\ref{subsubsec:CF-applyingTransformation}. The bibliographic references show matching observations found in the literature, when available.}
    \label{tab:SummaryPropertiesMeasures}
\end{table}

\subsubsection{Typology of Measures}
\label{sec:TypoMeasures}
We now show how a typology of the measures can be built based on the results shown in Figure~\ref{fig:RelativeImportanceResults}, through a cluster analysis. First, we compute a distance matrix comparing all pairs of stacked bars constituting the plots from this figure. For this purpose, we represent each stacked bar by a vector of proportions, each value corresponding to a term of the regression model (i.e. a segment of the stacked bar). We use the Hellinger distance~\cite{Cam1986}, which was designed to compare pairs of discrete probability distributions. Second, we perform the cluster analysis by applying the $k$-medoids method~\cite{Kaufman2009} to our distance matrix. This method requires us to specify the desired number of clusters, though. To find the most appropriate number, we apply the standard approach consisting in performing the clustering using all possible values, and then selecting the most appropriate one. For this purpose, we use the Silhouette measure, a well-known internal criterion~\cite{Rousseeuw1987}, but we also take into account a more subjective constraint of parsimony (i.e. we want a small number of clusters).

The analysis results in 5 clusters of stacked bars, for a Silhouette of $0.55$. Table~\ref{tab:PatternComparison} shows the distribution of the bars from Figure~\ref{fig:RelativeImportanceResults} over these clusters, each one being represented as a specific color. The blue cluster corresponds to bars in which there is a relatively balanced main effect of $k$ and $q$, and a minor effect of $h$ and $k$:$q$. In the brown cluster, the situation is quite similar but $q$ supersedes $k$. In the red cluster, $q$ even more prevalent, and both minor effects are even smaller. The orange cluster contains bars in which all effects are negligible compared to $q$. Finally, bars from the green cluster are dominated by $q$ and exhibit a minor effect of $h$:$q$.

\begin{table}[H]
\captionsetup{width=0.9\textwidth}
    \footnotesize
	\centering 
	\begin{tabular}{l p{2.4cm} p{2.4cm} p{2.4cm} p{2.3cm} p{2.3cm}}
    	\hline
    	 & \textbf{Orthogonal} & \textbf{k New} & \textbf{Neighbor} & \textbf{1 New} & \textbf{Singleton} \\
    	 & \textbf{Clusters} & \textbf{Clusters} & \textbf{Cluster Swaps} & \textbf{Cluster} & \textbf{Clusters} \\
    	\hline
        $D_{RI}$ & \cellcolor{MYBLUE} & \cellcolor{MYBLUE} & \cellcolor{MYBLUE} & \cellcolor{MYRED} & \cellcolor{MYBLUE}  \\
        $D_{ARI}$ & \cellcolor{MYRED} & \cellcolor{MYBROWN} & \cellcolor{MYBROWN} & \cellcolor{MYRED} & \cellcolor{MYRED}  \\
        $D_{FMI}$ & \cellcolor{MYORANGE} & \cellcolor{MYORANGE} & \cellcolor{MYGREEN} & \cellcolor{MYBROWN} & \cellcolor{MYORANGE} \\
        $D_{JI}$ & \cellcolor{MYORANGE} & \cellcolor{MYORANGE} & \cellcolor{MYGREEN} & \cellcolor{MYBROWN} & \cellcolor{MYORANGE} \\
        $D_{F}$ & \cellcolor{MYRED} & \cellcolor{MYORANGE} & \cellcolor{MYGREEN} & \cellcolor{MYBROWN} & \cellcolor{MYORANGE} \\
        $D_{NMI}$ & \cellcolor{MYRED} & \cellcolor{MYBLUE} & \cellcolor{MYBLUE} & \cellcolor{MYRED} & \cellcolor{MYBROWN} \\
    	\hline
	\end{tabular}
    \caption{Comparison of the measures based on the characterization provided by our framework and shown in Figure~\ref{fig:RelativeImportanceResults}. We use the Hellinger distance and $k$-medoids to identify groups of similar behaviors, each one being represented by a color in the table.}
    \label{tab:PatternComparison}
\end{table}

Table~\ref{tab:PatternComparison} shows that each transformation produces a different vertical pattern, which indicates that the transformations we selected in our framework are not redundant in the way they allow characterizing the measures. The measures can be compared using the horizontal patterns present in the table. Roughly speaking, there is a first group constituted of $D_{FMI}$, $D_{JI}$, $D_{F}$; a second containing $D_{RI}$ and $D_{NMI}$; and $D_{ARI}$ is apart. We see that this characterization is consistent with the results in Table~\ref{tab:SummaryPropertiesMeasures}. The fact that these groups of measures, which are automatically obtained, match the ones identified manually based on our knowledge of the desired properties, indicates that this clustering-based method could be useful when the user is not able to (or does not want to) express their desired properties \textit{a priori}. Indeed, for a given collection of available measures, this method allows identifying clusters of measures possessing a similar behavior: these clusters can then be characterized \textit{a posteriori}, and the user can select a measure from the cluster considered as the most appropriate to the considered application.

To sum up, not only does our analysis allows distinguishing the effects of the framework parameters over transformation types and measures, but it also makes it possible to categorize the measures based on their empirical behavior. Our results confirm the findings of Pfitzner \textit{et al}.~\cite{Pfitzner2008}, which indicate that the categorization of the measures based on their sole definitions (cf. Section~\ref{subsec:Measures}) does not necessarily hold when it comes to comparing them through experiments.

\section{Practical cases}
\label{sec:PracticalCases}
In practice, an external evaluation measure is usually needed in two situations frequently occurring in the context of cluster analysis or community detection. In the first, one wants to compare an estimated partition to a partition of reference. This typically happens when one has applied some algorithm in order to estimate a partition of their data, and wants to quantify how similar it is to some available ground truth partition. In this context, the measure is used to assess the performance of the partitioning method. 
In the second situation, there is no reference partition involved: one wants to compare two estimated partitions. For instance, one has access to several partitions and wants to assess them in the absence of any ground truth. These partitions could either result from the application of several distinct partitioning methods to the same data, or from the application of single method able to output several solutions for the same input data. In this context, one would use a measure to assess how similar these partitions are, in order to check whether the methods reach a relative consensus. 

The external measure has a central role in both situations, as different measures are likely to result in very different outcomes. The choice of an appropriate measure depends on a number of factors, including the broad situation, but also the nature of the application at hand and other contextual aspects such as the behavior expected by the user. In particular, it is worth stressing that not all transformations and parameters are relevant in all cases.

In the following, we illustrate all these aspects through two use cases, each one corresponding to one of the two broad situations described above. First, we treat the partitioning of the well-known cluster analysis method $k$-means, in a case where the ground truth is known (Section~\ref{subsec:ComparingGroundTruthAndEstimatedPartitions}). Second, we turn to the cluster analysis of a set of estimated partitions, in the context of a study aiming at analyzing votes at the European Parliament (Section~\ref{subsec:ComparingEstimatedPartitions}).

\subsection{Comparing Estimated Partitions with Ground-Truth}
\label{subsec:ComparingGroundTruthAndEstimatedPartitions}
In order to illustrate the comparison of some estimated partitions with the ground truth, we leverage the work of Fränti \& Sieranoja. In~\cite{Fraenti2018}, they study the behavior of 
$k$-means, and more precisely how this clustering method is affected by certain properties of the considered data. To this end, they propose a benchmark constituted of various artificially generated datasets together with their associated ground truth. Certain properties of these data are controlled through a set of parameters. Among these, some are only related to the $k$-means algorithm (e.g. number dimensions of the data, spatial overlap) and not to the problem of partition comparison, so we ignore them in the rest of our discussion. Fränti \& Sieranoja want to assess how changes in the properties controlled by the parameters affect the algorithm performance. For this purpose, they use the ARI and the Centroid Index (CI). The latter is a clustering comparison measure defined by Fränti \textit{et al}. in a previous article~\cite{Fraenti2014}. It focuses on \textit{global} partition differences concerning the number of clusters, by opposition to what the authors call \textit{point-level differences}, i.e. \textit{local} differences concerning the cluster borders. However, it was designed specifically to handle centroid-based clustering methods, which is why it is not part of the selection of measures we study in Section~\ref{subsec:Measures}.

Let us now suppose that one wants to use a measure selected among the one discussed in Section~\ref{subsec:Measures}. We can leverage the objectives of Fränti \& Sieranoja as described in~\cite{Fraenti2018}, as well as their methodology, to infer which measure behavior is desirable in terms of our own framework. First, the main parameters used to \textit{directly} control the ground truth partitions in \cite{Fraenti2018} are the number of objects, which corresponds to our parameter $n$, and the number of clusters, which is the same as our $k$. Although it is not controlled by a specific parameter, they also consider datasets with various levels of cluster size imbalance, a feature related to our $h$. The authors compare scores produced on data obtained by using different values of these parameters. For these comparisons to be relevant, it is necessary that these parameters affect the measure as little as possible, in order for it to reflect only changes in algorithm performance. 

It appears clearly in the article that, for the authors, incorrectly estimating the number of clusters is the most serious error that $k$-means can make, by opposition to so-called point-level errors which concern only cluster borders. The transformations of our framework which are the most relevant to this situation are therefore those that change the number of clusters, i.e. all of them but \textit{Neighbor Cluster Swaps}. Moreover, due to the nature of the considered data and clustering method, it is very unlikely to see singleton clusters appear in the considered partition (this would require the presence of very eccentric outliers). Therefore, transformation \textit{Singleton clusters} is not relevant in this situation. This leaves us with \textit{1 New Cluster}, \textit{$k$ New Clusters} and \textit{Orthogonal Clusters}.

Let us now assess the relevance of the measures studied in Section~\ref{subsec:Measures} with respect to the criteria we identified. Based on our results from Section~\ref{subsec:ResultsRelativeImportance}, we can identify two categories of measures in this situation. First, $D_{RI}$, $D_{ARI}$ and $D_{NMI}$ are sensitive to $k$, especially for \textit{$k$ New Clusters} and \textit{Orthogonal Clusters}, and to a lesser extent, to $h$. They differ on \textit{1 New Cluster}, as $D_{RI}$ is much less sensitive to these parameters when considering this transformation. The second category contains $D_{FMI}$, $D_{JI}$ and $D_{F}$, which exhibit sensitiveness to $k$ and $h$ only for \textit{1 New Cluster}, and \textit{Orthogonal Clusters} in the case of $D_{F}$. In conclusion, the second category is more appropriate to the situation described in~\cite{Fraenti2018}, with a preference for $D_{JI}$ which, overall, is less sensitive to the parameters of interest than the $D_{ARI}$ used in the original study.

\subsection{Comparing Estimated Partitions}
\label{subsec:ComparingEstimatedPartitions}
We now illustrate the case of comparing several estimated partitions with each other. In~\cite{Arinik2020}, Arinik \textit{et al}. study voting data from the European Parliament (EP), in order to identify voting patterns, i.e. how Members of the EP (MEP) are split in various factions depending on the topic of the considered legislative texts. Formally, they model the MEPs' voting behavior as a multiplex signed graph, and perform community detection on each layer to identify so-called voting patterns, i.e. partitions of the set of MEPs. This specific application brings specific constraints on the partitions, which can contain at most three communities: 1) a single community in case of unanimity (all MEPs vote either \textit{For} or \textit{Against} the legislative text); 2) two communities when there is either an antagonistic situation (i.e. some MEPs support the concerned document and the rest oppose it), or a unanimous community with an additional community of abstentionists; 3) two antagonistic communities with an additional community of abstentionists.

Arinik \textit{et al}. identify the partition of MEPs (voting pattern) associated to each legislative text in their corpus. They then use an external measure to compute the dissimilarity between each pair of partitions, and perform a cluster analysis in order to identify groups of similar patterns, which they finally discuss relatively to the application context. In order to select the most appropriate measure, they adopt a qualitative approach consisting in identifying some partitions of interest and comparing how different measures behave when comparing them. Among $D_{RI}$, $D_{F}$, $D_{ARI}$ and $D_{NMI}$, they conclude that $D_{F}$ and $D_{RI}$ are the most appropriate for their situation, with a slight advantage to $D_{F}$.

We propose to use our results from Section~\ref{subsec:ResultsRelativeImportance}, and in particular from Figure~\ref{fig:RelativeImportanceResults}, to solve the same measure selection problem, but based on the method presented in this article. Note that, in the following, we use the term \textit{cluster} instead of \textit{community}, for the sake of consistency with the rest of the article. It is important to stress that some parameters and transformations are not relevant here, due to the application context.
First, the case of $k=1$ is not applicable for some transformations. Therefore, we apply all our transformations if there is more than one cluster in the original partition. For \textit{k New Clusters} and \textit{Singleton Clusters}, this means that we get at least four clusters in the transformed partition. This is incompatible with the fact that all the compared partitions of this application contain at most three clusters, so we exclude both transformations. Second, in this context, the \textit{Orthogonal Clusters} transformation can be applied only when there are two clusters in the original partition, and only one element in each cluster is affected by the transformation. In this case, this transformation results in the same transformed partition as with \textit{1 New Cluster}, therefore we also exclude \textit{Orthogonal Clusters}. 

This leaves us with two transformations. The first is \textit{1 New Cluster}, which 
we apply only when the original partition has two clusters, in which case the transformation produces an additional cluster in the transformed partition. The second is \textit{Neighbor Cluster Swaps}, which 
we apply only when the original partition has either two or three clusters, but not a single one. For both these transformations, there is no constraint on parameters $h$ and $q$. However, as explained above, our analysis must focus only on certain values of $k$. Finally, in this context all the considered partitions contains the same number of elements, which means $n$ is fixed and can therefore be ignored in our discussion.

Next, based on the description given by Arinik \textit{et al}. of what they consider to be an appropriate measure for their application needs, we express the desired behavior of the measure with respect to the remaining parameters and transformations. 
The measure must be sensitive to the \textit{1 New Cluster} transformation as, in this context, detecting an extra cluster or missing one is an important error, since there are only a few possible clusters of MEPs. When $k$ increases, so does the diversity of the cluster created by this transformation, in the sense that its elements come from more distinct original clusters. In this context, this is an important difference with the original partition, so we want the score of the measure to increase with $k$. By comparison, it is desirable that the dissimilarity score decreases when $h$ increases, as this means most elements of the extra cluster come from the same original cluster, an error which is less serious. For the same reason, the effect of $k$ should be stronger than that of $h$.
Transformation \textit{Neighbor Cluster Swaps} consisting in mixing the original clusters to get the transformed partition without changing the number of clusters makes the clusters more different, when $q$ increases. In this application context, it is important that this type of difference between partitions is taken into account, so the measure must be sensitive to it. Changes in $k$ and $h$ do not affect the mixing much, so the measure score is expected to be largely independent from these parameters.

Let us now study which measures studied in Section~\ref{subsec:Measures} fit the constraints described above. Regarding transformation \textit{1 New Cluster}, it appears that only $D_{FMI}$, $D_{JI}$ and $D_{F}$ behave appropriately. When considering transformation \textit{Neighbor Cluster Swaps}, we can see that, even if it is a small one, $k$ as an effect on $D_{FMI}$ and $D_{JI}$. In conclusion, based on these observations, we would select $D_{F}$ in this context, a choice that incidentally matches the one made through a more qualitative and heuristic method in~\cite{Arinik2020}.


\section{Conclusion}
\label{ref:Conclusion}
In this article, we have presented a new evaluation framework to address the problem of selecting an appropriate measure to compare partitions. We want not only to compare measures, but also to produce results that the end user can easily interpret. For this purpose, based on our review of the literature, we designed a set of predefined partitions and parametric partition transformations in order to generate a benchmark dataset. Our two-step framework first computes the considered measures for these partitions, then conducts a regression and relative importance analysis to determine how the measures are affected by the transformations. We illustrated its relevance by applying it to a selection of standard measures.
We showed that our framework allows identifying the desirable properties possessed by each measure. For some of them, our results confirm empirical and theoretical findings already published in the literature. For others, the systematic nature of our approach even uncovers properties not mentioned before in the literature.
Furthermore, we propose a typology of the considered measures based on their characteristics. Overall, our results confirm the findings of Pfitzner \textit{et al}.~\cite{Pfitzner2008}, which indicate that categorizing measures based on their mathematical definitions does not necessarily match experimental comparison. Finally, we demonstrated how our framework can be put in practice through two concrete use cases: comparing an estimated partition to a partition of reference, and comparing several estimated partitions with each other.

Our work could be extended in several ways. First, our method can be applied systematically to other external measures, for the sake of completeness. It is particularly important to include the recently proposed measures for an up-to-date comparison, which would prevent from following the tradition of using only well-established measures without regard for their relevance. Second, similar to the previous point, some new parametric transformations can be proposed to closely investigate the performance of the measures on a specific subject. For instance, there is an important number of measures aiming at correcting Mutual Information for chance in the literature. Including some specific transformations could enable to concentrate more on the aspect related to the number of clusters. Finally, by proposing relevant parameters and transformations, our general method could be adapted to handle objects similar to partitions, such as covers, to compare overlapping clusters (e.g. \cite{Horta2015, Gates2017b}), or edge-aware community similarity measures, to compare community structures while taking graph topology into account (e.g. \cite{Rabbany2013, Labatut2015}).
 

\section*{Acknowledgment}
The authors thank Thomas Opitz, Etienne Klein from INRAE PACA and Pierre-Michel Bousquet from LIA for their feedback and guidance on certain statistical points.

\phantomsection\addcontentsline{toc}{section}{References}
\printbibliography
%

\appendix

\section{Additional Results}
\label{appendix:AddtionalResults}

\begin{figure*}[ht!]
\captionsetup{width=0.9\textwidth}
	\centering
    \includegraphics[width=0.8\linewidth]{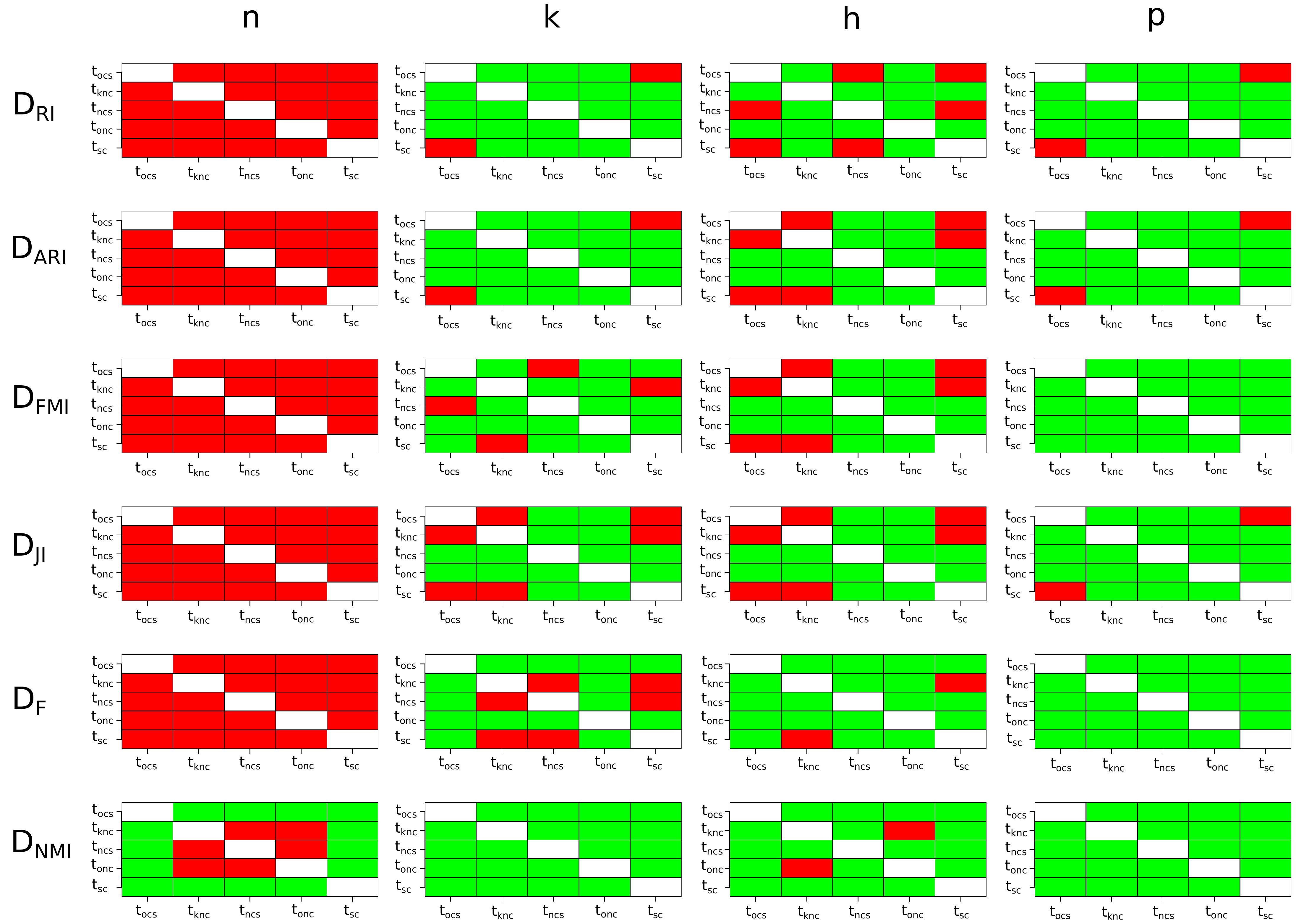}
	\caption{Significance of the results regarding the comparison of the segment heights performed in Section~\ref{subsec:ResultsRelativeImportance} over all pairs of transformations, considered for each measure and parameter set. For instance, the top four matrices correspond to Figure~\ref{fig:RelativeImportanceResults}.a. Green (resp. red) cells represent significant (resp. non-significant) differences between the considered transformations, with a significance level of $\alpha = 0.05$. Figure available at \href{https://doi.org/10.6084/m9.figshare.13109813}{10.6084/m9.figshare.13109813} under CC-BY license.}
	\label{fig:Appendix_significance_diffrences_for_measures}
\end{figure*}

\begin{figure*}[ht!]
\captionsetup{width=0.9\textwidth}
	\centering
    \includegraphics[width=0.9\linewidth]{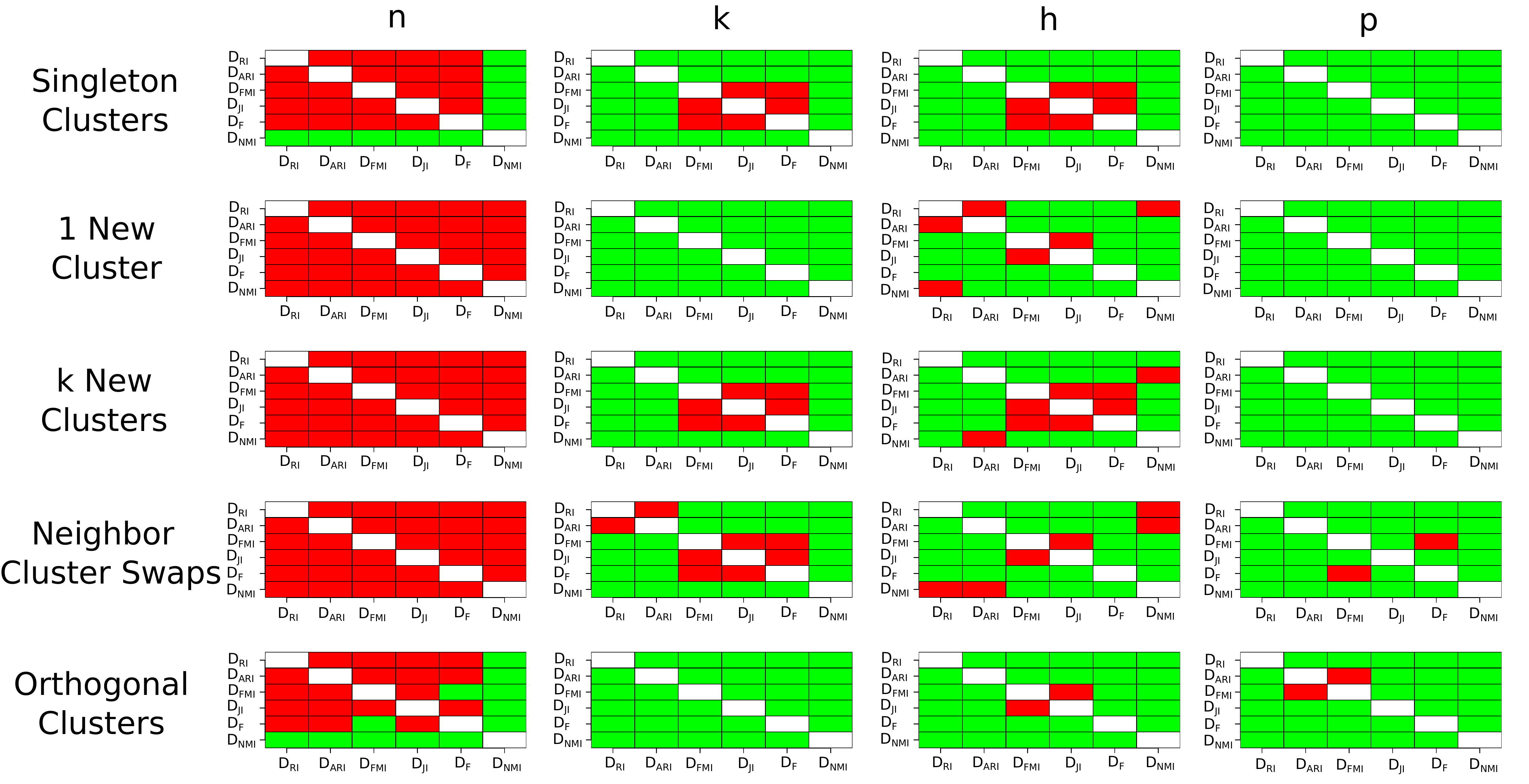}
	\caption{Significance of the results regarding the comparison of the segment heights performed in Section~\ref{subsec:ResultsRelativeImportance} over all pairs of measures, considered for each transformation and parameter set. For instance, the top four matrices correspond to the last stacked bar in each barplot of Figure~\ref{fig:RelativeImportanceResults} (\textit{Singleton Clusters}). Green (resp. red) cells represent significant (resp. non-significant) differences between the considered measures, with a significance level of $\alpha = 0.05$. Figure available at \href{https://doi.org/10.6084/m9.figshare.13109813}{10.6084/m9.figshare.13109813} under CC-BY license.}
	\label{fig:Appendix_significance_diffrences_for_transfs}
\end{figure*}

\section{Evaluation Measures}
\label{appendix:EvalMeasures}
In this section, we give the formal definition of the evaluation measures used in this work. A common point of those measures is that they can be computed using the so-called \textit{confusion matrix} (also called \textit{association matrix} or \textit{contingency table}) based on the two partitions. 

We note $n$ the numbers of elements of a dataset $D$. Also, let $P = \{C_1,...,C_k\}$ ($1 \leq k \leq n$) be a $k$-partition of $D$, i.e. a division of $D$ into $k$ non-overlapping and non-empty clusters $C_i$ ($1 \leq i \leq k$). Let have another partition $P'$ formed by $k'$ clusters, where $k'$ may be different from $k$. Then, the \textit{confusion matrix} is a $k\ \times\ k'$ integer matrix, whose $ii'$th cell is the number of elements in the intersection of clusters $C_{i}$ and $C_{i'}$, as shown in Table~\ref{tab:ContingencyTable}.

\begin{table}[H]
\captionsetup{width=0.9\textwidth}
	\caption{The confusion matrix for two partitions $P = \{C_1,...,C_k\}$ and $P' = \{C'_1,...,C'_{k'}\}$ of $n$ elements, where $n_{ij} = |C_i \cap C'_j|$ are the number of elements in both clusters $C_i \in P$ and $C'_i \in P'$.}
    \label{tab:ContingencyTable}
	\centering
	\begin{tabular}{r c | c c c | c }
      				& 		     &  &  Partition $P'$ &  & \\
        			& Cluster    & $C'_{1}$ & $\dots$ & $C'_{k'}$ & Marginal sum\\
        \hline
         			& $C_{1}$    & $n_{11}$ & $\dots$ & $n_{1k'}$ & $n_{1\cdot}$ \\
     Partition      & $\cdot$    &    $\cdot$     &     &    $\cdot$      &    $\cdot$     \\
    $P$ 			& $\cdot$    &    $\cdot$     &     &    $\cdot$      &    $\cdot$     \\
         			& $C_{k}$  & $n_{k1}$ & $\dots$ & $n_{kk'}$ & $n_{k\cdot}$ \\
        \hline
      				& Marginal sum 	    & $n_{\cdot1}$ & $\dots$ & $n_{\cdot k'}$ & $n_{\cdot\cdot} = n$\\
	\end{tabular}
\end{table}

\subsection{Rand Index, \textit{RI}}
The formulation of all pair-counting measures can be expressed in terms of four types of element pairs. The \textit{positive agreement} $N_{11}$ corresponds to the number of element pairs which are in the \textit{same} cluster in \textit{both} partitions $P$ and $P'$. The \textit{negative agreement} $N_{00}$ is the number of element pairs which are in \textit{different} clusters in \textit{both} $P$ and $P'$. The partitions \textit{disagree} on the remaining element pairs, as $N_{10}$ (resp. $N_{01}$) corresponds to the number of element pairs which are in the same cluster in $P$ (resp. $P'$), but not in $P'$ (resp. $P$). The formula of each term is shown in Table~\ref{tab:contingency-table-pair-types}.

\begin{table}[h]
	\caption{Formulae for the number of (unordered) element pairs of the four types}
    \label{tab:contingency-table-pair-types}
	\centering
	\begin{tabular}{l l}
        \hline
      	Type 	& 	Formula	\\
        \hline
        $N_{11}$	&	$\displaystyle \sum\limits_{i=1}^{k}\sum\limits_{j=1}^{k'} \binom{n_{ij}}{2} = \frac{\displaystyle 1}{\displaystyle 2} \sum\limits_{i=1}^{k}\sum\limits_{j=1}^{k'} n_{ij}(n_{ij} - 1)$  \\
        $N_{00}$	&   $\displaystyle \binom{n}{2} - (N_{11} + N_{10} + N_{01})$ \\
        $N_{10}$	&	$\displaystyle \frac{\displaystyle 1}{\displaystyle 2}\Big( \sum\limits_{i=1}^{k}  n_{i \cdot}^2 - \sum\limits_{i=1}^{k}\sum\limits_{j=1}^{k'}n_{ij}^2 \Big)$ \\
        $N_{01}$	&	$\displaystyle \frac{\displaystyle 1}{\displaystyle 2}\Big( \sum\limits_{j=1}^{k'} n_{\cdot j}^2 - \sum\limits_{i=1}^{k}\sum\limits_{j=1}^{k'}n_{ij}^2 \Big)$ \\
        \hline
        $N_{\cdot \cdot} = N_{11} ~~+ N_{10} + N_{01} + N_{00}$ & $\displaystyle \binom{n}{2} = n(n-1)/2$ \\
        \hline
	\end{tabular}
\end{table}

The \textit{Rand Index} (RI)~\cite{Rand1971} is the proportion of total agreement, i.e. when counting both positive and negative agreement:
\begin{equation}
	RI(P, P') = \frac{N_{11} + N_{00}}{N_{\cdot \cdot}}.
    \label{eq:Rand}
\end{equation}
Its values lie between $0$ and $1$, where $0$ occurs for the absence of any positive and negative agreements, whereas $1$ corresponds to the case where the partitions are perfectly identical.

\subsection{Adjusted Rand Index, \textit{ARI}}
The \textit{Adjusted Rand Index} (ARI)~\cite{Hubert1985} is a well-known extension of the Rand Index, with additional correction for chance. It aims at dealing with the statistical independence of two partitions  (see Section~\ref{subsubsec:HandlingIndependentPartitions}). Its formula is
\begin{equation}
	ARI(P, P') = \frac{RI(P,P') - \mathop{\mathbb{E}}[RI(P, P')]}{1 - \mathop{\mathbb{E}}[RI(P, P')]}.
	\label{eq:AdjustedRand}
\end{equation}
where $\mathop{\mathbb{E}}[RI(P, P')]$ corresponds to the estimated score of $RI(P, P')$ for independent partitions under hypergeometric assumption (so-called permutation model). This term is defined as
\begin{equation}
\mathop{\mathbb{E}}(\sum\limits_{ij}\binom{n_{ij}}{2}) = \sum\limits_{i}\binom{n_{i \cdot}}{2} \sum\limits_{j}\binom{n_{\cdot j}}{2} / \binom{n}{2}.
\end{equation}

The ARI takes a value of $1$ for identical partitions, whereas $0$ indicates a case of statistical independence. Moreover, ARI can take a negative value for very dissimilar partitions~\cite{Meila2015}, when the observed RI is smaller than expected.


\subsection{Jaccard Index, \textit{JI}}
The Jaccard Index (JI) was originally defined to compare sets~\cite{Jaccard1901}, but it is also used as an external measure~\cite{Ben-Hur2001}. As reported in \cite{Meila2015}, the negative agreement $N_{00}$ can be often almost as large as the maximum number of element pairs $\binom{n}{2}$. The Jaccard Index is an improved version of RI on this aspect, as it does not take $N_{00}$ into account. It is defined as
\begin{equation}
	JI(P, P') = \frac{N_{11}}{N_{11} + N_{01} + N_{10}}.
	\label{eq:Jaccard}
\end{equation}

The Jaccard Index ranges from $0$ (absence of any positive agreement) to $1$ (identical partitions). Note that one minus the Jaccard Index is a metric on the finite sets~\cite{Marczewski1958}.

\subsection{Fowlkes-Mallows Index, \textit{FMI}}
The Fowlkes-Mallows Index~\cite{Fowlkes1983} is the final pair-counting measure that we consider in this work. It was originally introduced to ease the comparison of hierarchical dendrograms. Like the Jaccard Index, it ignores negative agreements. It can be described as the geometric mean of two asymmetric forms of positive agreement: the proportion of positive agreements relative to the number of pairs belonging to the same cluster in $P$ vs. those in $P'$. Its formal description is
\begin{equation}
	FM(P, P') = \frac{N_{11}}{\sqrt{(N_{11} + N_{10}) (N_{11} + N_{01})}}.
	\label{eq:FowlkesMallows}
\end{equation}

\subsection{F-measure, \textit{F}}

In the category of set-matching measures, we select the $F$-measure (F). Note that this name is sometimes used in the literature as a synonym of \textit{harmonic mean}, and therefore covers several distinct measures (e.g.~\cite{Rabbany2013, Gates2017}). We use the definition of Artiles \textit{et al}.~\cite{Artiles2007}, according to which the $F$-measure is the harmonic mean of two quantities called \textit{Purity} and \textit{Inverse Purity}.


The formal definition of \textit{Purity} is as follows:
\begin{equation}
	Purity(P,P') = \sum\limits_{i}\frac{n_{i \cdot}}{n} \max_{j} \frac{n_{ij}}{n_{i \cdot}}.
\end{equation}	

The Inverse Purity is simply the Purity of the second partition relative the first, i.e. $Purity(P',P)$. 
Finally, the $F$-measure is the harmonic mean of the Purity and Inverse Purity
\begin{equation}
    F(P,P') = 2 \frac{Purity(P,P') \times Purity(P',P)}{Purity(P,P') + Purity(P',P)}.
    \label{eq:FMeasure}
\end{equation}

\subsection{Normalized Mutual Information, \textit{NMI}}
The last measure that we consider is the Normalized Mutual Information (NMI), which belongs to the category of information-theoretical measures. It is based on the notions of \textit{entropy} and \textit{Mutual Information}~\cite{Cover2006}. The principle behind these notions is to consider each partition as a categorical random variable, whose possible values are the clusters.

In the context of clustering, entropy in the sense of Shannon is defined as
\begin{equation}
	H(P) = -\sum\limits_{i=1}^{k} \frac{n_{i \cdot}}{n}\log \frac{n_{i \cdot}}{n}.
	\label{eq:Entropy}
\end{equation}

Each element in dataset $D$ has an equal probability of being picked, so its probability of being in cluster $C_{i}$ is $n_{i}/n$. Thus, we have a discrete random variable taking $k$ values, which is associated to the partition $P$. If the partition $P$ has only $1$ cluster containing all the points, then $H(P)$ will be zero, since there is no uncertainty in the clustering structure. If the partition $P$ consists of as many clusters as $n$, it will reach its maximum value. Note that $H(P)$ does not depend on $n$, but on the relative proportions of the clusters. 

The Mutual Information can be described as the mutual dependence between these variables, and it can then be interpreted as the similarity between the partitions. It is formally described as
\begin{equation}
    MI(P,P') = \sum\limits_{i=1}^{k}\sum\limits_{j=1}^{k'} \frac{n_{ij}}{n}\log\frac{\frac{n_{ij}}{n}}{\frac{n_{i \cdot}}{n} \frac{n_{\cdot j}}{n}}.
	\label{eq:MutualInformation}
\end{equation}

There are a number of variants of the notion of mutual information, in particular several normalizations have been proposed (see for instance~\cite{Vinh2010}). In this work, we focus on the sum normalization as defined in~\cite{Kvalseth1987, Strehl2002}, which is very widespread. The resulting NMI is
\begin{equation}
    NMI(P,P') = \frac{2 MI}{H(P)+H(P')}.	
\end{equation}


\section{Experimental details about the heterogeneity of cluster sizes}
\label{appendix:DetailsOfClusterSizeHeterogeneity}
There are many ways to make clusters imbalanced. In this work, we opt for a sequence based on an arithmetic progression. Consider the sizes of the clusters in a partition as a sequence of values $S_k$ whose sum is equal to the number of nodes $n$, as in \eqref{eq:Appendix_Equation_Sk}. 
In this equation, $\alpha$ corresponds to the first value and $\beta$ corresponds to the constant increment value
\begin{equation}
\label{eq:Appendix_Equation_Sk}
\begin{split}
    n & = \alpha + (\alpha+\beta) + (\alpha+2\beta) + ... + (\alpha+(k-1) \beta)\\
    & = \alpha k + \frac{\beta k (k-1)}{2}.
\end{split}
\end{equation}
Note that the sequence contains as many terms as the number of clusters.

In such a sequence, each term is a constant increment value $\beta$ larger than the previous term (e.g. $\beta=2$ for the sequence 3, 5, 7, ..). This $\beta$ is computed based on the parameter $h$ (heterogeneity of cluster sizes). When $h=1$, it reaches its maximal value that we note $\beta_{max}$. In the case of $h<1$, $\beta$ is proportional to $\beta_{max}$ to the extent of $h$ (i.e. $\beta = h \times \beta_{max}$). The value of $\beta_{max}$ can be determined in different ways. In order not to introduce an additional parameter for this, our approach is to assign the first term $\alpha$ and the constant increment $\beta_{max}$ to the same value, i.e. $\alpha = \beta_{max}$. Then, we compute $\beta$ as follows
\begin{equation}
    \label{eq:Appendix_Equation_Beta_max}
    \begin{split}
    n & = \frac{\beta_{max} k (k+1)}{2} \\
    \beta & = \floor{h \beta_{max}}.
    \end{split}
\end{equation}

Note that $\floor{.}$ denotes the floor function (returning the greatest integer less than or equal to the input value). Finally, we obtain the value of $\alpha$ as follows
\begin{equation}
    \label{eq:Appendix_Alpha}
    \begin{split}
    \alpha & = \frac{n - \frac{\beta k (k-1)}{2}}{k}.
    \end{split}
\end{equation}
 

\end{document}